\documentclass[journal]{IEEEtran}
\IEEEoverridecommandlockouts

\usepackage{lineno,microtype,subcaption}
\usepackage[onehalfspacing]{setspace}
\usepackage{soul}
\usepackage{orcidlink}
\usepackage{graphicx}
\usepackage{amsmath,amssymb,amsfonts,amsthm}
\usepackage{mathtools}
\usepackage{xcolor}
\usepackage{booktabs}
\usepackage{multirow}
\usepackage{tabularx}
\usepackage{array,makecell}
\usepackage{arydshln}
\usepackage{algorithm}
\usepackage{algpseudocode} % (algorithmicx base)
\usepackage{listings}
\usepackage[most]{tcolorbox}
\sethlcolor{yellow}
\usepackage{dblfloatfix} 
\title{Intent-Driven LLM Ensemble Planning for Flexible Multi-Robot Disassembly: Demonstration on EV Batteries
\vspace{-0.3em}}

\author{
\IEEEauthorblockN{%
Cansu Erdogan\textsuperscript{\dag,1}\orcidlink{0000-0002-8317-0237},
Cesar Alan Contreras\textsuperscript{\dag,1}\orcidlink{0009-0005-2866-8672},
Alireza Rastegarpanah\textsuperscript{\dag,1,2,*}\orcidlink{0000-0003-4264-6857},
Manolis Chiou\textsuperscript{3}\orcidlink{0000-0002-9779-4067},
Rustam Stolkin\textsuperscript{1}}
\IEEEauthorblockA{\textsuperscript{1}Extreme Robotics Lab, School of Metallurgy \& Materials, University of Birmingham, Birmingham, UK}
\IEEEauthorblockA{\textsuperscript{2}Department of Applied Artificial Intelligence and Robotics, School of Computer Science, Aston University, Birmingham, UK}
\IEEEauthorblockA{\textsuperscript{3}School of Electronic Engineering and Computer Science, Queen Mary University of London, London, UK}
\thanks{\textsuperscript{\dag}\,Co-first authors. * Corresponding author: a.rastegarpanah@aston.ac.uk.}
\thanks{This work is funded by the project called “Research and Development of a Highly Automated and Safe Streamlined Process for Increasing Lithium-ion Battery Repurposing and Recycling” (REBELION) under Grant 101104241, and partially supported by the Ministry of National Education, Republic of Turkey.}
\vspace{-2.5em}
}

\begin{document}

\maketitle

% ---------------------------------------
% Abstract
% ---------------------------------------
\begin{abstract}
This paper addresses the problem of planning complex manipulation tasks, in which multiple robots with different end-effectors and capabilities, informed by computer vision, must plan and execute concatenated sequences of actions on a variety of objects that can appear in arbitrary positions and configurations in unstructured scenes. We propose an intent-driven planning pipeline which can robustly construct such action sequences with varying degrees of supervisory input from a human using simple language instructions. The pipeline integrates: (i) perception-to-text scene encoding, (ii) an ensemble of large language models (LLMs) that generate candidate removal sequences based on the operator's intent, (iii) an LLM-based verifier that enforces formatting and precedence constraints, and (iv) a deterministic consistency filter that rejects hallucinated objects. The pipeline is evaluated on an example task in which two robot arms work collaboratively to dismantle an Electric Vehicle battery for recycling applications. A variety of components must be grasped and removed in specific sequences, determined by human instructions and/or by task-order feasibility decisions made by the autonomous system. On 200 real scenes with 600 operator prompts across five component classes, we used metrics of full-sequence correctness and next-task correctness to evaluate and compare five LLM-based planners (including ablation analyses of pipeline components). We also evaluated the LLM-based human interface in terms of time to execution and NASA TLX with human participant experiments. Results indicate that our ensemble-with-verification approach reliably maps operator intent to safe, executable multi-robot plans while maintaining low user effort.
\end{abstract}

\begin{IEEEkeywords}
Large Language Models, Multi-Robot Disassembly, Task Planning, Intent Recognition, Human-Robot Interaction
\vspace{-1.4em}
\end{IEEEkeywords}

% ---------------------------------------
\section{Introduction}
\vspace{-0.3em}
Traditional pre-programmed automation has worked well in highly structured settings, becoming widespread in automotive manufacturing and other factory assembly tasks since the 1970s. However, unstructured, variable, and uncertain environments pose additional challenges which are a bottle-neck to robotization of many other societally important industries.
The disassembly of electric-vehicle (EV) batteries is a prototypical task that demands strong generalisation from multi-robot systems. Battery packs vary widely across makes, models, and model years; there is no stable standard, and designs evolve continuously, so the classic ``re-program on every new variant'' approach is untenable \cite{hathaway2024technoeconomic}. Instead, scalable automation must (i) use computer vision to robustly recognise and localise heterogeneous components and infer how they fit together, (ii) plan low-level actions (grasping, moving, unscrewing, cutting) for specific robots on specific parts under perception uncertainty, (iii) schedule and coordinate multi-robot, multi-action task sequences, and (iv) provide an intuitive, high-level interface so a recycling worker (without robotics expertise) can quickly instruct the system. Similar needs arise in other unstructured settings (e.g. sorting and separating mixed waste streams for recycling; remote decommissioning of contaminated legacy facilities in high-hazard environments). Traditional Task and Motion Planning (TAMP) pipelines, which rely on handcrafted rules and rigid schedules, struggle to handle such variability \cite{asif2024robotic}, motivating the intent-driven, perception-grounded approach we pursue in this paper.

\begin{figure*}[!hb]
    \centering
    \includegraphics[width=0.80\linewidth]{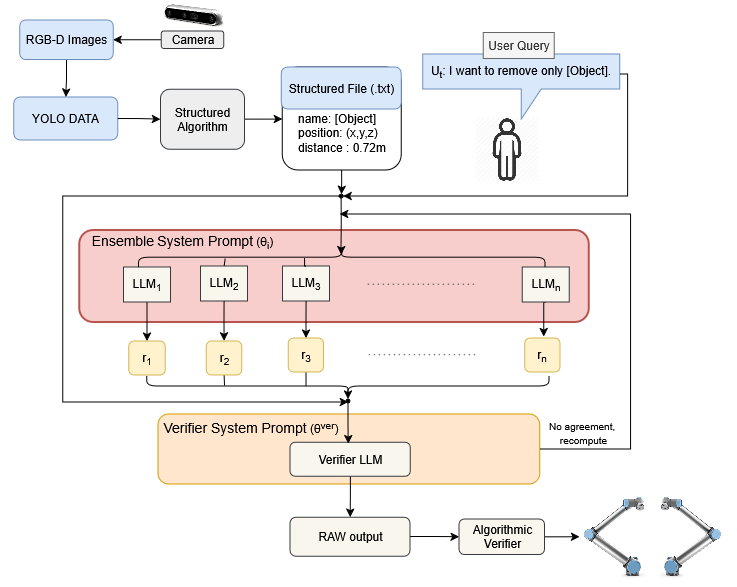}
    \caption{A task-planning pipeline that enhances reliability and flexibility by using an ensemble of LLMs (same checkpoint, different seeds) to generate diverse solutions from structured inputs. These candidate plans are then synthesised and validated by a dedicated verification module before robot execution.}
    \label{fig:final_pipeline}
\end{figure*}

Recent advances in large-scale pre-trained language models (LLMs) offer new opportunities for flexible, high-level reasoning over symbolic and spatial descriptions. By leveraging LLMs’ ability to process and synthesise contextual information, it becomes feasible to generate executable multi-robot plans from textual scene descriptions combined with operator intent. Coupled with perception for perceptual grounding of task representations and motion planning for execution, this capability reduces manual modelling effort and enables a more adaptive, scalable planning pipeline.

\subsection{Overview}
This study introduces a multi-LLM-based task Planning framework for multi-robot disassembly operations. A user writes what they want taken apart, and the system turns that request into a step-by-step plan that two robots can safely carry out. More specifically, the approach integrates vision-based scene understanding modules with an LLM-driven reasoning layer to infer object ordering and produce abstract task plans. These are then grounded into executable sequences via a motion-level planner, which enforces physical constraints and robot coordination. We conduct a structured evaluation based on a custom dataset of battery disassembly scenarios, using metrics such as task ordering accuracy and plan execution success rates. The final pipeline is shown in Figure \ref{fig:final_pipeline}.

%\vspace{-1em}

This work makes four key contributions: (i) an end-to-end, language-conditioned disassembly pipeline which maps operator intent and live perception to executable, precedence-constrained action lists for multi-arm systems and robot platforms with heterogeneous capabilities; (ii) a domain-agnostic perception–language interface which serializes detections from text-output vision models (YOLOv8) as structured text with coordinates in the operator frame, enabling spatial reasoning while decoupling detection from planning; (iii) a planner which uses a single instruction-tuned checkpoint (Qwen3-32B) with stochastic sampling to produce diverse candidates, followed by an LLM verifier and a deterministic consistency filter to enforce precedence and format, and reject objects that are ungrounded; and (iv) a workload-aware human-in-the-loop execution layer which coordinates multiple manipulators, using a digital twin for collision checks and synchronized trajectory streaming, providing a simplified hand-off from plan to execution to reduce operator cognitive and interaction workload.

\section{Literature Review}\label{sec2} 
Our setup comprises multiple components, including task planning, large language models, ensemble methods, operator intent, and adaptive autonomy. This section reviews the relevant work for each component and motivates the design choices used later.
\vspace{-0.1em}

\subsection{Task planning and robot allocation} 
In multi-robot systems, task planning involves selecting and sequencing a set of high-level actions that collectively achieve a system-level objective. At the same time, robot allocation assigns these actions to individual agents based on capabilities, resources, availability, and spatial constraints. Coordination between both layers is necessary for ensuring efficient and collision-free movements in shared environments. Classical optimisation methods are often used, such as Mixed-Integer Linear Programming (MILP) and Constraint Programming (CP), that encode task assignment under resource and precedence constraints \cite{FATEMIANARAKI2023102770,9681247}. These approaches are very effective in structured industrial scenarios. Still, they may suffer from scalability issues and degrade their performance in dynamic or uncertain environments.

To improve scalability and adaptation, recent works have explored the use of heuristic and distributed scheduling methods \cite{WANG2024104604, 10802695}, which show more responsive and decentralised decision-making capabilities under uncertainties. Knowledge-based and semantic reasoning frameworks \cite{BERNARDO2023109345} have introduced domain semantics awareness into task planning, to interpret object roles, workspace conditions, and task dependencies. Despite progress, many frameworks still rely on handcrafted rules, limiting their adaptability to new tasks or environments. These limits motivate data-driven approaches that can adapt with less manual modelling.

We replace much of the handcrafted task-ordering logic with an LLM ensemble that infers precedence-consistent removal sequences from general disassembly instructions, a text-based scene encoding, and an operator prompt. We then double-check these sequences with an LLM verifier and a deterministic data-consistency filter before any motion is executed. The framework also lets the operator edit the plans and take control at any time.

\subsection{Large Language Models on robotics}
The integration of LLMs into robotics has opened avenues for high-level task planning, semantic perception, and human-robot interaction. Among early impactful frameworks, RoboLLM demonstrated strong performance on vision-language benchmarks such as ARMBench, showing how LLMs can be grounded in visual inputs for robotic manipulation tasks \cite{long2024robollmroboticvisiontasks}. However, these benchmarks typically represent static or highly structured environments. Real-world multi-robot systems tend to operate in unstructured and uncertain conditions.

Working under these constraints, the 3D-LOTUS++ framework in Towards Generalizable Vision-Language Robotic Manipulation combines the high-level reasoning of LLMs with the spatial awareness of VLMs for object localisation. This integration improves generalisation in robotic manipulation \cite{garcia2025generalizablevisionlanguageroboticmanipulation}. Another framework working in this direction is ManipLLM, which introduces a multimodal LLM-based manipulation system that focuses on object-centric reasoning \cite{li2023manipllmembodiedmultimodallarge}. Unlike earlier learning-based methods limited to predefined object categories, ManipLLM employs commonsense reasoning chains and test-time adaptation strategies to generalise across diverse manipulation tasks, performing well in simulated and real-world environments.

For human–robot interaction (HRI), recent work has emphasised collaboration with LLMs, including studies of LLM-enabled robots performing execution, negotiation, selection, and plan generation \cite{Kim_2024}, as well as vision-language-action reasoning combined with teleoperation \cite{Liu_2024}. Findings suggest that while LLMs excel in affective and negotiation-based interactions, they struggle to maintain logical consistency and to support creative collaboration.

In our work we operationalise LLM planning for disassembly tasks by (i) converting RGB-D detections into a structured text scene, (ii) sampling multiple plans from a single LLM checkpoint with different seeds, (iii) verifying logical ordering and precedence with an LLM instructed for verification tasks, and (iv) applying a final algorithmic data-consistency filter, to reduce hallucinations, before commanding the robots.

\subsection{Ensembles and Instruction Tuning}
Instruction tuning (IT) adapts a pretrained language model to follow natural-language prompts by fine-tuning it on instruction triples (instruction-input-output). The supervised fine-tuning (SFT) pipeline begins with a training stage over dozens to hundreds of different tasks \cite{wei2022finetunedlanguagemodelszeroshot}, and sometimes is followed by a reinforcement learning step with human feedback (RLHF) using their preferences data \cite{ouyang2022traininglanguagemodelsfollow}. Exposing the systems to very diverse instructional data and human feedback gives strong zero-shot generalisation on unseen tasks, and also helps reduce toxic behaviours \cite{wei2022finetunedlanguagemodelszeroshot, ouyang2022traininglanguagemodelsfollow, chung2022scalinginstructionfinetunedlanguagemodels}. Early evidence of the strength of SFT came from FLAN, which fine-tuned a 137B parameter model on 60+ different tasks and surpassed bigger models in zero-shot accuracy \cite{wei2022finetunedlanguagemodelszeroshot} with subsequent works showing that scaling the number of tasks and the model parameters improves instruction-following abilities, and that injecting chain-of-thought traces during fine-tuning also boosts reasoning accuracy \cite{chung2022scalinginstructionfinetunedlanguagemodels}.

The increase of IT models prompted the design of specialised benchmarks that evaluate models on their obedience to explicit constraints, with some of these including:  \textbf{IFEval}, which judges whether generated text satisfies verifiable instructions (e.g length or lexical constraints) \cite{zhou2023instructionfollowingevaluationlargelanguage}, \textbf{FollowBench} dissects adherence across content, format, and style dimensions \cite{jiang2024followbenchmultilevelfinegrainedconstraints} and \textbf{M-IFEval}, which adds judgement of multilingual instructions, comparing it to the default IFEval \cite{dussolle2025mifevalmultilingualinstructionfollowingevaluation}.

Ensemble methods can complement Instruction Tuning.  Deep ensembles improve calibration by aggregating independently sampled models \cite{lakshminarayanan2017simplescalablepredictiveuncertainty}. Model-library selection ensembles pick a diverse subset from thousands of candidates for optimal accuracy \cite{caruana2004ensemble}.  Sparse mixture-of-expert (MoE) architectures, using the Switch Transformer, scale to trillions of parameters and some obtain ensemble-like gains fusing final layers at fixed compute by routing a subset of tokens through selected layers \cite{fedus2022switchtransformersscalingtrillion, raposo2024mixtureofdepthsdynamicallyallocatingcompute}. Ensembles of instruction-tuned generators in theory could produce higher-quality synthetic training data, and self-consistency voting across multiple IT models, reducing reasoning errors at inference. Yet despite the progress in each direction, studies on ensemble design choices tailored to instruction-following constraints are still limited.

In our work, instead of extra fine-tuning, we perform ensemble-at-inference on a single instruction tuned LLM checkpoint (In a 1/3/6 ensemble of Qwen3-32B), and add an LLM verifier for format, precedence and logical planning verification. We also enforce an algorithmic filter that checks for hallucinations, as this process is faster and more adaptable than retraining.

\subsection{Operator Intent}
The ability to understand and respond to an operator’s intentions is a prerequisite for fluent human-robot teaming.  Early Bayesian formulations from motions showed that predicting a collaborator’s next goal allows a robot to adapt its plan, yielding improvements in both objective and perceived performance \cite{liu2018goalinferenceimprovesobjective}. Some works have also framed intent recognition as an online classification problem, in teleoperation work, for example,  navigation and manipulation intents are able to be inferred and classified with onboard sensing, robot trajectories, saliency and geometric cues \cite{TSAGKOURNIS20238333, contreras2025probabilistic, panagopoulos2021bayesian} reaching real-time accuracy comparable to hand-engineered baselines. 

Vision-only, zero-shot intent recognition has also shown reductions in operator effort on unseen manipulation tasks \cite{belsare2025zeroshotuserintentrecognition}. In multi-robot settings, inferring collective operator intent introduces additional operator and bandwidth challenges, with some researchers on human-swarm interaction marking cognitive load as the main bottleneck when conveying intent to dozens of agents simultaneously \cite{7299280}. Shared-control systems take it one step further by blending user commands with autonomy in proportion to the certainty of inferred intent, and sometimes intent-aware assistance is preferred over full autonomy  \cite{Bowman_2024}.

Taking from the literature the importance of assistance and intent recognition, in this work we treat natural-language prompts as the explicit operator intent and fuse them with the scene text so the ensemble orders only disassembly orders that satisfy the prompt (e.g. "remove only leafcell"), while the verifier enforces logical disassembly precedence and the filter prevents objects not present in the scene from appearing, giving this calculations to the system keeps cognitive-load low, while it also preserves trust by allowing operators to verify and change disassembly order before execution.

\subsection{Variable Autonomy}
Variable autonomy (VA) refers to a robot's ability to modulate its level of independence, shifting decisions and control between human operators and on-board autonomy in response to context, workload, mission phase, or robot conditions \cite{contreras2025mini, methnani2024s, chiou2021mixed, reinmund2024variable}. Some adjustable-autonomy frameworks formalise transfer of control strategies, balancing decision quality against miscoordination costs, and show performance gains for single operators supervising several robots at varying autonomy levels \cite{scerri2002towards, crandall2001experiments}. More recently, some research has framed VA as a prerequisite for trustworthy AI, emphasising transparency and explainability as determinants of when autonomy should be ceded or reclaimed \cite{methnani2024s}. 

Language-centric VA mechanisms in multi-robot tasks on VR testbeds have also been worked on \cite{Lakhnati_2024}, and have been shown to be helpful in deciding when to act autonomously and when to ask an operator for help, further expanding the available methods to do Variable Autonomy. Variable autonomy rules and allocation rules can also be learned, as an example, the ATA-HRL framework utilises hierarchical reinforcement learning to re-allocate tasks and determine autonomy mode in multi-human multi-robot teams, outperforming fixed-autonomy baselines under dynamic mission conditions \cite{yuan2025adaptivetaskallocationmultihuman}. 

Our planner supports on-the-fly autonomy adjustment via language; operators may change objectives mid-execution while the LLM verifier and algorithmic data-consistency filter preserve safety. A motion layer allows the coordination of multiple arms without restarting, allowing to keep a low cognitive workload, and the system also lets operators take control at any time.

\section{Method}\label{sec3}
\begin{figure}[!hb]
    \centering
    \includegraphics[width=0.9\linewidth]{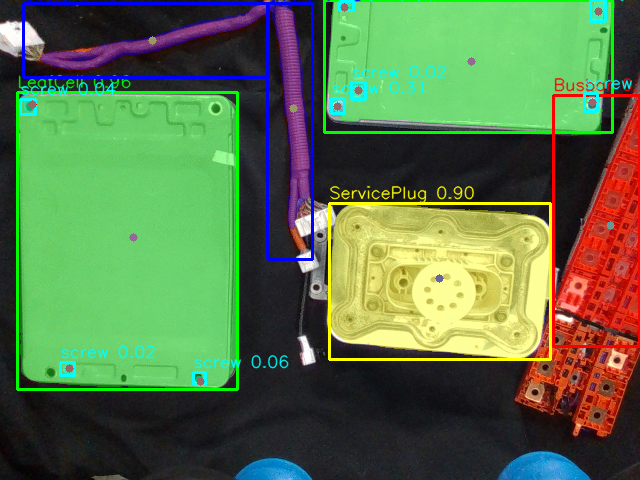}
    \caption{Example perception output used by the LLM-ensemble planner. A YOLOv8 model overlays class labels and confidence scores for the battery components (\textit{Leafcell}, \textit{Cable}, \textit{Busbar}, \textit{Service Plug}, \textit{Screw}).}
    \label{fig:SceneImage}
\end{figure}

When planning for robot disassembly, task hierarchies tend to be static. However, this limitation prevents systems from adapting to environmental changes. With vision models, it has been possible to adapt to changing positions; however, adapting to different operator intentions and explicit needs remains a challenge. To test our dynamically adaptive method, we build on our previous papers \cite{robotics13050075, shaarawy2025multirobotvisionbasedtaskmotion}, and have designed a robotic scene consisting of two UR10e robots with cameras mounted on their end effectors, with the task of identifying objects forming part of a dissassembly task (Fig. \ref{fig:SceneImage}), and correctly disassembling them. For this task, we used weights of YOLOv8 with 226 images, similarly to our previous paper \cite{shaarawy2025multirobotvisionbasedtaskmotion} components, in a real-world environment. In addition to the usage of the pre-trained Qwen-32B LLM, without any additional fine-tuning.

\begin{figure*}[!h]
    \centering
    \includegraphics[width=0.65\linewidth]{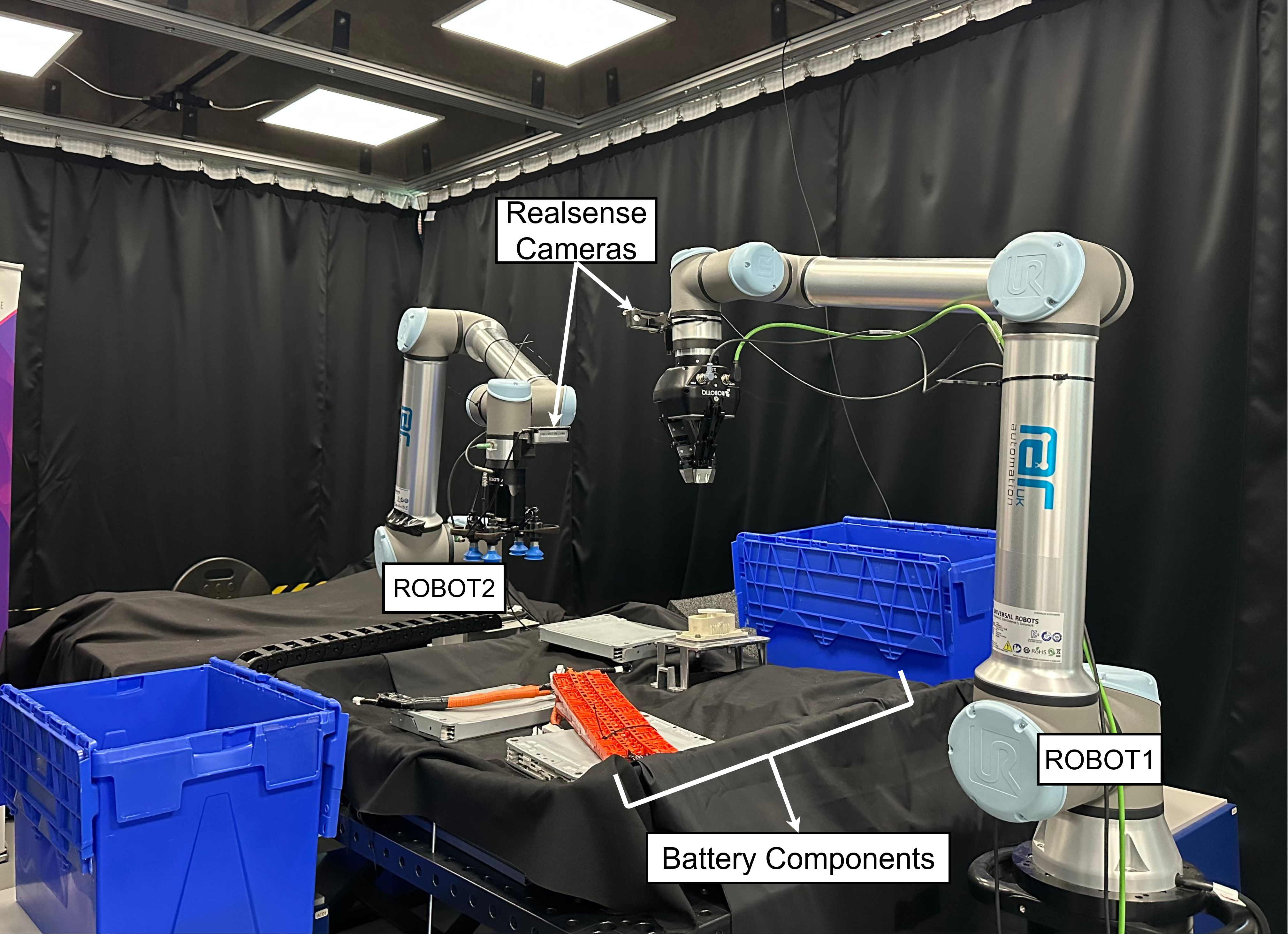}
    \caption{Data-collection setup: two UR10e collaborative robot arms (ROBOT1 and ROBOT2) equipped with Intel RealSense RGB-D cameras and the components arranged on the workbench.}
    \label{fig:text_analysis}
\end{figure*}

We deployed our experimental setup across two dedicated desktop computers. The first system, denoted as PC$_1$, was responsible for real-time perception and robotic control. PC$_1$ featured an Intel i7 CPU, an NVIDIA GTX 1080 Ti GPU, and 32\, GB of RAM. This machine was directly connected to the Intel RealSense 435i RGB-D cameras mounted on the robots (Fig. \ref{fig:text_analysis}), which continuously provided synchronised RGB and depth data. Additionally, PC$_1$ managed low-level control of two Universal Robots (UR-series) manipulators, handling joint commands and execution of planned motions.

The second system, designated PC$_2$, performed high-level reasoning tasks and hosted the language models utilised in our experiments. PC$_2$ was configured with an Intel i9 CPU, an NVIDIA RTX 5090 GPU, and 32\, GB of RAM. This computer exclusively ran a quantised (4-bit) version of the Qwen-32B large language model. All models were served through an Ollama-compatible API, and inter-system communication between PC$_1$ and PC$_2$ occurred over a WiFi network, with 200ms latency.

\subsection{Scene encoding} \label{sec:encoding_list}
Raw sensor state \(x_t\in\mathbb{R}^n\) is segmented by YOLOv8, yielding
\[
  s_t=g(x_t)=\{(b_k,c_k)\}_{k=1}^{K_t}.
  \label{eq:scene_pairs}
\]

The set \(\,s_t\) is rendered as an unstructured string and then structured and serialised \(\sigma_t=\text{Serialise}(s_t)\).

\begin{tcolorbox}[title=Excerpt of Unstructured Text \(s_t\),colback=gray!8,colframe=gray!40]
\footnotesize\ttfamily
THIS IS THE PUBLISHED DATA  (dist mode: XYZ)
LeafCell   pos[    1.304     0.877    -0.081] dist=1.574  bbox(m): TL[    1.454     0.873     0.048] TR[    1.455     0.884    -0.202] BR[    1.143     0.882    -0.204] BL[    1.141     0.870     0.046] C[    0.189     0.003     0.669]\\
LeafCell   pos[    1.309     0.854     0.186] dist=1.574  ...\\
Cable      pos[    1.516     0.900     0.059] dist=1.764  ...\\
screw      pos[    1.167     0.852     0.092] dist=1.448  ...\\
screw      pos[    1.438     0.846     0.270] dist=1.690  ...\\
screw      pos[    1.431     0.854     0.103] dist=1.669  ...\\
screw      pos[    1.169     0.866     0.032] dist=1.455  ...\\
screw      pos[    1.167     0.843     0.285] dist=1.467  ...\\
\end{tcolorbox}

\begin{tcolorbox}[title=Excerpt of Structured Text \(\sigma_t\),colback=gray!8,colframe=gray!40]
\footnotesize\ttfamily
Group 1: LeafCell is in location [1.304, 0.877, -0.081] with screw [1.169, 0.866, 0.032] on top of it.\\
Group 2: LeafCell is in location [1.309, 0.854, 0.186] with screw [1.167, 0.852, 0.092], screw [1.438, 0.846, 0.27], screw [1.431, 0.854, 0.103], screw [1.167, 0.843, 0.285] on top of it.\\
Group 3: Cable is in location [1.516, 0.9, 0.059].\\
\end{tcolorbox}

\subsection{Coordinate rewrite}
Every Cartesian triple \([x,y,z]\) inside \(\sigma_t\) is replaced by \([-z,\,x,\,y,\,d]\) with
\begin{equation}
  d=\sqrt{x^{2}+y^{2}+z^{2}},
  \label{eq:coord_norm}
\end{equation}
, due to the transformations from the camera to spatial positions on the operator frame, while all other tokens are left untouched.  
Write
\begin{equation}
  D_t=\mathcal T(\sigma_t)
  \label{eq:Dt}
\end{equation}
for the transformed scene text produced by the coordinate rewriting.

\subsection{Generator-system prompt}
The prompt \(\theta_i\) transforms the backbone model into a \textit{Generator}: from the structured scene description and the operator’s instruction, it must produce, without commentary, an ordered list of objects that must be removed.  To keep the generator and verifier perfectly aligned, the prompt first clarifies the role and output contract, then reminds the model of the coordinate convention already applied to the received input.  A compact input-grammar description follows, covering how leads, followers, and their positions appear in each \texttt{Group} statement.  The prompt then details a domain-agnostic target-resolution algorithm that defines how to locate requested objects, accumulate any mandatory blockers, and order the final removal set.  A series of self-checks ensures that the list is complete, free of duplicates, and respects all ordering rules before emission. Explicit error strings allow the downstream runner to detect fatal conditions early, while a disallowed-actions section forbids headings, blank lines, fabricated data, or any explanatory text.  Finally, a bank of illustrative examples anchors the model’s behaviour on different corner cases without revealing internal task labels.

\begin{tcolorbox}[title=Generator system prompt \(\theta_i\), colback=blue!5, colframe=blue!60]
\footnotesize\ttfamily
- Role declaration\\
- Coordinate convention\\
- Input grammar description\\
- Target resolution algorithm\\
- Self-checklist \\
- Error strings\\
- Disallowed actions\\
- Illustrative examples
\end{tcolorbox}

\subsection{Generator pass}
Let \(u_t^{\text{extra}}\in\mathcal U\) be the operator instruction and \(\theta_i\in\Theta\) the static system prompt for model \(i\).
The user-role message is $U_t$

\begin{tcolorbox}[title=Final User Message \(U_t\), colback=blue!6,colframe=blue!45]
\footnotesize\ttfamily
Here is the data to use: $\langle D_t\rangle$\\
This is what I want to do 'extra prompt': $\langle u_t^{\text{extra}}\rangle$\\
/no\_think
\end{tcolorbox}

With sampling seed \(s_j\), the message sent to the LLM API is
\begin{equation}
  p_{i,j,t}=
  [(\text{"system"},\theta_i),\;
   (\text{"user"},U_t)].
  \label{eq:payload}
\end{equation}

\begin{tcolorbox}[title=Generator LLM Message, colback=blue!6,colframe=blue!45]
\footnotesize\ttfamily
"system", $\theta_i$,\\
"user", $U_t$.
\end{tcolorbox}

\subsection{LLM candidate generation}
Let \(\mathcal S\) be the (finite) set of sampling seeds for the single backbone model \(\varphi_i\). For every \(s_j\in\mathcal S\) the model processes the payload  \(p_{i,j,t}\) from \eqref{eq:payload} and returns a chat trace
\begin{equation}
  c_{i,j,t}= \varphi_i(p_{i,j,t}).
  \label{eq:raw_trace}
\end{equation}
Define the function \(\operatorname{\textit{last}}:\text{trace}\rightarrow\text{string}\) that keeps only the final bullet list of a trace.
The seed-specific candidate text is therefore
\begin{equation}
  r_{i,j,t}= \operatorname{last}(c_{i,j,t}).
  \label{eq:cand_text}
\end{equation}

\subsection{Verifier-system prompt logic}
The verifier system prompt \(\theta^{\mathrm{ver}}\) guides the LLM into a verification-focused role, independent from the previous generation step. Its purpose is to enforce consistency and correctness among candidate responses. The prompt is explicitly structured into functional sections, starting with a clear role definition, followed by a description of the structured input layout provided by the user: this includes the initial context (system and user prompts) and candidate responses produced earlier. 

A concise summary of the reference algorithm then follows, outlining how ground truth should be recomputed from the structured input without relying on task-specific terminology. Subsequently, the prompt defines multiple categories of validation criteria, including strict formatting requirements to ensure syntactic uniformity, presence constraints that mandate exact matches within the context, ordering rules to maintain algorithmic consistency, and a policy to eliminate duplicate entries. After validation, the verifier is tasked with employing a simple decision procedure, selecting the first candidate list that fulfils all specified criteria. If no candidate meets all conditions, the verifier must explicitly return an empty output, prohibiting any form of merging, partial acceptance, or speculative editing. 

Finally, the prompt concludes by precisely specifying the required output format and explicitly restricting actions that fall outside its scope, such as commentary, partial lists, or explanations of additional context. This design systematically ensures deterministic and reliable verification without disclosing or embedding unnecessary domain knowledge, enabling it to function with different unspecified object labels.
\vspace{-0.4em}
\begin{tcolorbox}[title=Verifier system prompt \(\theta^{\mathrm{ver}}\), colback=green!6,colframe=green!45]
\footnotesize\ttfamily
- Role and responsibilities definition\\
- Structured input description\\
- Generalised algorithm recap\\
- Validation criteria\\
- Decision and selection logic\\
- Explicit output formatting rules\\
- Restricted actions and prohibitions
\end{tcolorbox}

\subsection{Verifier pass}
Let
\[
  C_t = \operatorname{str}(\theta_i,\,U_t)
\]
be the printable pair comprising the generation system prompt \(\theta_i\) and user message \(U_t\).
Denote by
\(
\mathcal R_t=\{\,r_{j,t}\mid s_j\in\mathcal S\}
\)
The set of seed-specific candidate lists harvested in
\eqref{eq:cand_text}.  
A fixed deterministic formatter
\(\Phi:\bigl(\text{context},\text{set of strings}\bigr)\to\text{string}\)
produces the user-side input to the verifier:
\[
  \xi_t=\Phi\!\bigl(C_t,\mathcal R_t\bigr).
\]
The same LLM checkpoint is invoked once more, now under a dedicated
verifier system prompt \(\theta^{\mathrm{ver}}\); its output is
\[
  v_t = g\!\bigl(\theta^{\mathrm{ver}},\xi_t\bigr),
\]
A single cleaned bullet list in plain text.

\begin{tcolorbox}[title=Verifier LLM Message, colback=green!6,colframe=green!45]
\footnotesize\ttfamily
system:\;$\theta^{\mathrm{ver}}$\\
user:\\
\quad=== ORIGINAL CONTEXT ===\\
\quad$\langle C_t\rangle$\\[4pt]
\quad=== CANDIDATE RESPONSES ===\\
\quad Response 1:\\
\quad $\langle r_{1,t}\rangle$\\[2pt]
\quad Response 2:\\
\quad $\langle r_{2,t}\rangle$\\
\quad $\vdots$\\
\quad Response $|\mathcal S|$:\\
\quad $\langle r_{|\mathcal S|,t}\rangle$\\
\quad/no\_think
\end{tcolorbox}

\subsection{Coordinate extraction}
A deterministic parser \(\psi\) strips every
\texttt{\string<think\string>} block from the verifier output \(v_t\)
and converts each item into a pair
\((\hat\ell^{(n)},\hat{\mathbf q}^{(n)})\) where
\(\hat{\mathbf q}^{(n)}=(L,U,Z)^{\mathsf T}\).
The object coordinates are then mapped back to the original frame via
\((L,U,Z)\mapsto(x,y,z)=(U,Z,-L)\), yielding
\[
  \hat c_t
  =\psi(v_t)
  =\bigl\{(\hat\ell^{(n)},\hat{\mathbf q}^{(n)})\bigr\}_{n=1}^{\hat M_t}.
\label{eq:ver_output}
\]

\subsection{Algorithmic deterministic verifier/filter}
A pair is retained when its text appears verbatim in the original scene text \(\sigma_t\) (this to remove hallucinations that the ensemble of generators or the verifier might introduce).
\[
  \chi(\hat\ell,\hat{\mathbf q};\sigma_t)=
  \begin{cases}
    1,& \text{if } \operatorname{regex}(\hat\ell,\hat{\mathbf q})
       \subset \sigma_t,\\[4pt]
    0,& \text{otherwise}.
  \end{cases}
  \label{eq:chi}
\]
The accepted list is therefore
\[
  A_t
  =\bigl\{(\hat\ell^{(n)},\hat{\mathbf q}^{(n)})
          \mid
          \chi(\hat\ell^{(n)},\hat{\mathbf q}^{(n)};\sigma_t)=1
   \bigr\}.
  \label{eq:accepted}
\]

\subsection{Robot action execution}
The high-level object lists generated by the LLM-ensemble are executed by a single \texttt{robot\_control} ROS node that commands both UR10e arms (ROBOT1\ \&\ ROBOT2). Built on MoveIt, this node performs motion planning, trajectory streaming, and run-time safety checks in one process, thereby eliminating the latency and synchronisation issues that arise when an independent planner drives each arm.\\

\textbf{Control pipeline,}  
\begin{itemize}
  \item \emph{Scene initialisation} - At start-up, the node constructs a shared MoveIt planning scene that includes both physical arms and a kinematic “digital twin’’ of the opposite arm to enable local inter-robot collision checks.  
  \item \emph{Task ingestion} - Verified object lines (e.g.\ \texttt{leafcell [x y z]}) arrive on a ROS topic. They are expanded into \textit{approach-grasp-retreat-place} primitives whose goal poses are supplied by the YOLO/RealSense perception stream. 
  \item \emph{Controller type(s) and Planning pipeline(s)} - velocity controller OMPL \cite{sucan2012the-open-motion-planning-library}, velocity limits: 20\% of maximum.\\
\end{itemize} 
For further implementation details, please refer to \cite{shaarawy2025multirobotvisionbasedtaskmotion}.

\subsection{Adjustable autonomy modes}\label{sec:va_modes}
The final control interface supports variable autonomy during planning by allowing the operator to modulate the plan source at runtime. \textbf{Full order (human-planned).} The operator specifies a complete ordered list; the verifier and filter check it, and the motion layer executes it. \textbf{Partial order (human goal, system completes).} The operator names a target object or a subset of steps. The LLM infers a precedence-consistent order and inserts mandatory prerequisites. \textbf{No order (system-planned).} The operator states only a goal (e.g., "remove the left leafcell"); the LLM proposes a full plan subject to verification and filtering. \textbf{Mid-execution override.} The operator may interrupt an autonomous sequence and issue a new intent; the pipeline replans without restarting the system.

These modes map control between human and autonomy depending on operator preference and task context, which defines the variable autonomy used in this work. 

\section{Experiment}\label{sec4}

To utilise the framework, a series of experiments was designed to prepare, test, and verify its functionality on our disassembly setup. The goal of the first set of experiments was to select a Language Model that accurately and efficiently followed instructions. The goal of the second set of experiments was to verify that the framework, with the selected LLM, generalises across a wide variety of conditions, contexts, and instructions. The goal for the third and final set of experiments was to verify its usability in the real world by conducting a pilot study with non-expert participants.

To perform this set of experiments, we generated a dataset that worked with our previously trained weights for YOLOv8 \cite{YOLOv8}. The collected dataset consisted of 200 RGB-D images, obtained with a RealSense 435i stereo camera. All RGB images passed through our YOLOv8 weights, which provided five different classes (screw, leafcell, cable, busbar and service plug)

\begin{figure}[!h]
    \centering
    \includegraphics[width=0.80\linewidth]{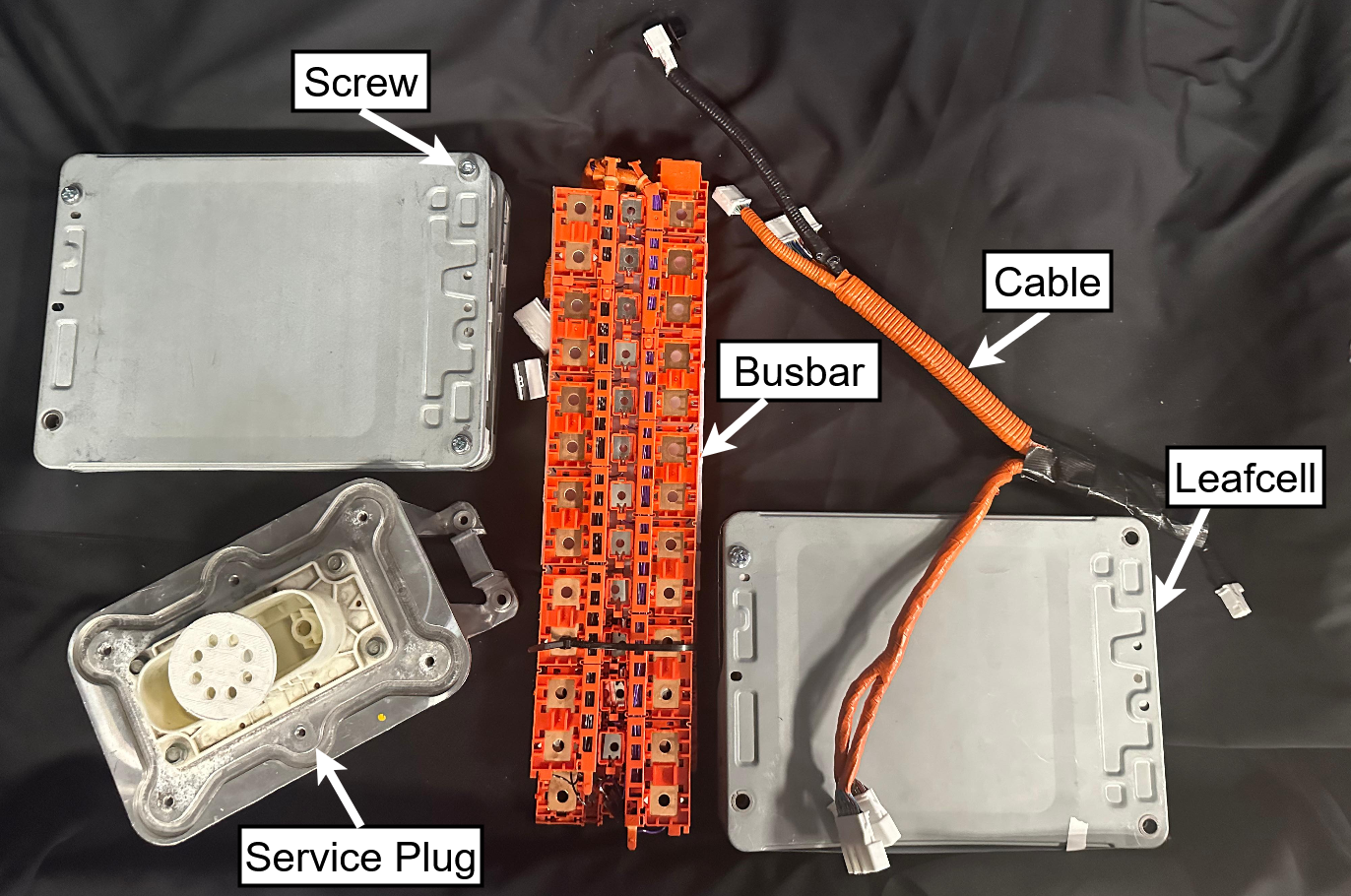}
    \caption{Sample scene showing five battery components (Leafcell, Busbar, Cable, Service Plug and Screw)}
    \label{fig:image_analysis}
\end{figure}

\begin{figure}[!h]
    \centering
    \includegraphics[width=0.70\linewidth]{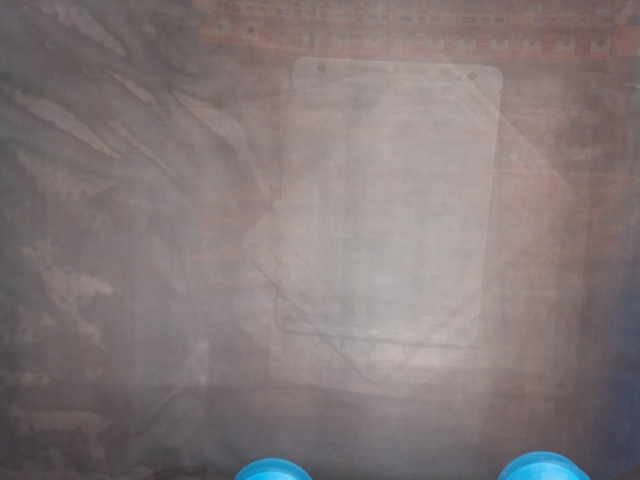}
    \caption{Ghost composite of all RGB frames. Frames are stacked and intensity-normalised. High-intensity regions indicate frequent occupancy across scenes; faint structures show sparse placements.}
    \label{fig:ghost_image}
\end{figure}

Figure \ref{fig:text_analysis} and Figure \ref{fig:image_analysis} show an example of object distribution and robot placement in the scene. Additionally, we analysed the object placement distribution by stacking every RGB frame of all trials, creating a blend of the distribution of the objects. Figure \ref{fig:ghost_image} illustrates the resulting composite, revealing where objects appeared throughout the experiment. The composite provides a spatial prior over the workspace: bright areas correspond to positions of components that appear often in the dataset; dim traces surface infrequent positions of components that would otherwise vanish in a simple average.

\begin{table*}[!hb]
\small
\caption{Screening results for candidate language models (five trials per model).
         All models were prompted with an identical instruction set.}
\label{tab:model_selection}
\begin{tabularx}{\textwidth}{l c X}
\toprule
\textbf{Model} & \textbf{Failures} & \textbf{Observation} \\
\midrule
\textbf{qwen3:32b [Reasoning Off]} &
0 &
Completed every trial; median latency approximately 21s (range 18-24s).  Response format is always correct, but inference speed is moderate compared with smaller models. \\

phi4:latest &
5 &
Failed to produce a valid plan in every trial, giving unnecessary explanations, even though runtimes stayed within 30-46s.  Errors stem from systematic instruction following issues rather than timeouts. \\

phi4-reasoning:latest &
5 &
No valid output across all runs.  Latency was highly unstable (16s-144s), indicating internal difficulties in reasoning steps, and overthinking of instructions. \\

phi4-reasoning:plus &
3 &
Succeeded in the first two trials (approximately 39-48s) but failed the remaining three, yielding a 40\% success rate.  Failures had many varied formatting issues in the output, although speeds were relatively fast. \\

gemma3:12b-it-qat &
5 &
Returned malformed responses in every trial with extra verbosity (approximately 127-129s). No successful completions observed. \\

gemma3:27b-it-qat &
3 &
Two successful runs (latencies 33s and 70s) but three failures caused by output format and verbosity errors, showing problems in instruction following steps; overall 40\% success and high variance in turnaround time. \\

deepseek-r1:latest &
5 &
Exceeded the 180s cutoff in all five trials; treated as hard failures with no usable output. \\

qwen3:30b-a3b &
3 &
Completed two trials (in approximately 36s) but failed three.  Failures produced verbose, off-specification answers in under 10s. \\
\bottomrule
\end{tabularx}
\end{table*}

\subsection{Selection of Language Model}\label{sec:sel_model}
Reliable plan generation demands an LLM that (i) follows our specialised instruction-example prompt format without hallucinations and (ii) produces answers quickly. It must also handle underspecified user input: operators may request any order as long as the text includes at least one object in view; the LLM must then infer a feasible, precedence-consistent order and insert any mandatory prerequisites (e.g., remove screws or cables before extracting a leafcell) that the user did not mention. We screened eight open-weight LLM models offline (Table \ref{tab:model_selection}). 

Each model received the identical user prompt $U_t$ and system prompt $\theta_i$. We ran five independent trials per model. A run failed if the output violated instructions (missing list, wrong delimiter, fabricated objects, verbosity that broke the format) or exceeded 180s. We logged completion time for successful runs.

The model \textbf{qwen3:32b} completed all five trials, though with a moderate median latency of 21s.  Three models, \textbf{phi4-reasoning: plus}, \textbf{qwen3:30b-a3b}, and \textbf{gemma3:27b-it-qat}, succeeded in roughly 40\% of runs but generated verbose or partially incorrect output in the remainder.  All other checkpoints failed every trial, either timing out or ignoring mandatory constraints.  Detailed outcomes appear in Table \ref{tab:model_selection}.

Given these results, we selected \textbf{qwen3:32b} for all subsequent experiments and demonstrations, prioritising accuracy and overall task completion rate.

\subsection{Pipeline for full correctness}\label{sec:ablation}

To identify the most effective pipeline configuration, a comparative evaluation was conducted across several distinct pipeline setups. Each variant employed the same underlying models and input prompts, differing only in their structural arrangement. The evaluated configurations included 1) a \textbf{single-model baseline}, 2) an \textbf{ensemble of three models combined with a verifier}, 3) an \textbf{ensemble of six models combined with a verifier}, 4) an \textbf{ensemble of three models coupled with a verifier and an additional algorithmic verifier step}, and finally, 5) an \textbf{ensemble of six models integrated with a verifier and the same algorithmic verifier}.

The single-model baseline consisted solely of the primary model (qwen3:32b), which executed directly using the provided prompts without any additional validation steps. Ensemble configurations involved running multiple models concurrently, obtaining multiple candidate responses. 

Subsequently, in ensemble setups, a verifier LLM took the candidate responses, systematically evaluated them according to a provided verification prompt, and returned a single, verified output. The ensemble variants marked \textit{with Algorithmic Verifier} added a final stage where the verified responses underwent an additional data consistency check. This step retained only responses explicitly matching the original input data, effectively removing any potentially hallucinated or incorrect information.

\begin{figure}[h]
    \centering
    \includegraphics[width=1\linewidth]{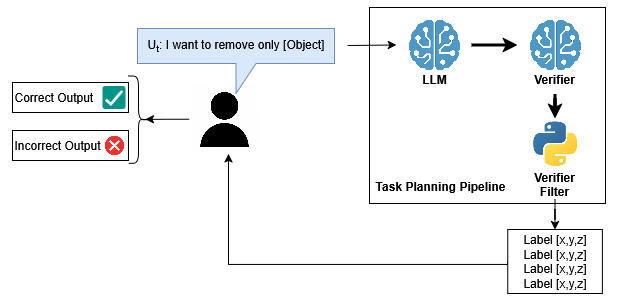}
    \caption{LLM + verifier architecture distinguishing correct ($\checkmark$) and incorrect ($\times$) output for full evaluation}
    \label{fig:all_objects}
\end{figure}

All evaluations were conducted using 600 user-generated prompts across 200 distinct real robot scenarios (3 per scenario). The responses were evaluated by operators, who determined whether the output met the logical requirements of the task while adhering to its given intent. Figure \ref{fig:all_objects} shows an example of an expected output from a "full" trial, where an ordered list of objects is likely to be the outcome. 

The data from the scenarios and each user prompt were recorded, and the tasks were repeated three times for each of the evaluated configurations. Each output was then assessed under consistent conditions and prompts, ensuring fairness and comparability across pipeline variants. The primary metrics recorded included the correctness of outputs, execution time per pipeline configuration, and adherence to instruction sets. In total, 9000 experiments were conducted across the five configurations, with the hypothesis that the model with the largest ensemble and both verifiers would yield higher accuracy than the single-model configuration.

\subsection{Pipeline for Next Object Correctness}\label{sec:ablation_next_object}

\begin{figure}[h]
    \centering
    \includegraphics[width=1\linewidth]{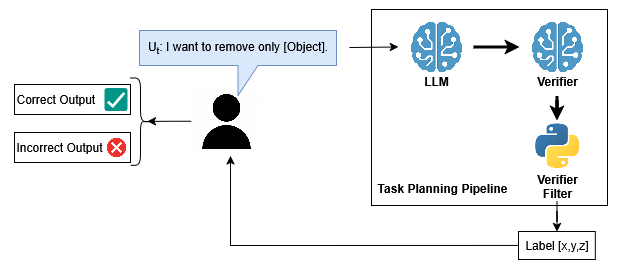}
    \caption{ LLM + verifier architecture distinguishing correct ($\checkmark$) and incorrect ($\times$) output for the first object}
    \label{fig:first_object}
\end{figure}

To further evaluate the effectiveness of each pipeline configuration, we performed a secondary analysis on the previously obtained experimental data. In this evaluation, we assessed only whether the first object in each generated response correctly matched the next object to be removed, according to the task logic and intent. The goal of this targeted evaluation was to determine if ensemble and verifier mechanisms specifically improve the initial selection within generated action plans. The verification flow for first-object accuracy is summarised in Figure \ref{fig:first_object}.

The same five pipeline configurations as in the "full" correctness trials were compared. No new experiments were conducted; instead, previously recorded outputs from the original 9000 trials (600 user-generated prompts across 200 scenarios, repeated three times per configuration) were re-evaluated against this narrower correctness criterion. Figure \ref{fig:all_objects} shows an example of an expected output from a "next object correctness" trial, where from the ordered list, only the first object is kept in the output.

The primary metric recorded was the accuracy of the first suggested object in the response, specifically assessing whether it matched the correct object designated for immediate removal. This evaluation enabled a direct comparison of pipeline configurations to determine whether the ensemble and verifier steps provided a clear advantage over the single-model baseline, particularly in terms of the accuracy of initial task predictions.

\subsection{Real-world Usability Experiment}\label{sec:real_world_demo}

We conducted a real-world usability pilot experiment to assess how non-experts issue intent-driven disassembly commands using our LLM-ensemble pipeline. Seven volunteers (\textit{n}=7), with no prior experience controlling robots, interacted with the system via a dedicated front end. The goal was to evaluate operator user experience and workload when commanding multi-robot disassembly via natural language.

Each participant issued two intents (one per arm) in plain language. The pipeline then generated an ordered list per arm that respected the disassembly precedence (without the need to give all objects of the intended disassembly explicitly). 

After a brief, single training session covering scene representation, UI elements, and camera views, participants were instructed to operate a two-arm disassembly system by specifying \emph{what} to remove or disassemble (one natural-language intent per arm). They saw two example intents during training (e.g., “remove only the leafcell from the left module”), then wrote their own intents for each scene. The study evaluated a single condition: the full pipeline (selected backbone LLM with the 6 LLM ensemble, verifier, and final consistency filter) integrated with two UR10e arms; no alternative planners or interfaces were tested here (comparisons of the pipeline appear in the offline object correctness experiments), but a 300s baseline of an expert operator performing the task without LLM assistance is recorded. 

Participants reviewed each object appearing on the Camera Views and in the list of possible objects (screw, leafcell, cable, busbar, and service plug) before taking any action. 

The procedure for each trial experiment was: (1) press \textit{Capture} to run YOLOv8 and generate the structured text scene and 3D view; (2) type one intent per arm; (3) press \textit{Let LLM plan} to obtain the removal lists; (4) verify that the lists matched the desired human intent and press \textit{Go} to execute; (5) after execution, report whether each action reflected the true intent and complete the NASA-TLX questionnaire. The primary measures were time-to-action and perceived workload (NASA-TLX); secondary measures included plan acceptance rate and execution success.

For each participant, we recorded: (i) \textit{LLM plan success} (Yes/No) for each arm based on participant confirmation of their intention; (ii) \textit{time-to-action} (seconds until pressing "Go" with a correct plan); and (iii) \textit{NASA-TLX} workload questionnaire.

To contextualise time-to-action, we used a \textbf{manual baseline} from an expert operator. The operator ran YOLOv8, reviewed the detections, and hand-entered coordinates according to his own disassembly plan for three items from the YOLO position list. He then executed the plan (“Go”). The mean over three runs was 300 s. This baseline is not a paired control; it is only for interpreting the pipeline times.

After completing the NASA-TLX questionnaire, the participants were allowed to issue extra commands for exploration; these interactions were not logged as part of the study dataset. Participants informally reported that the system was very simple to use and that they could have completed the trials without the training session. Some noted that planning also worked with non-English requests during their exploration attempts.

\begin{figure*}[!hb]
    \centering
    \includegraphics[width=0.99\linewidth]{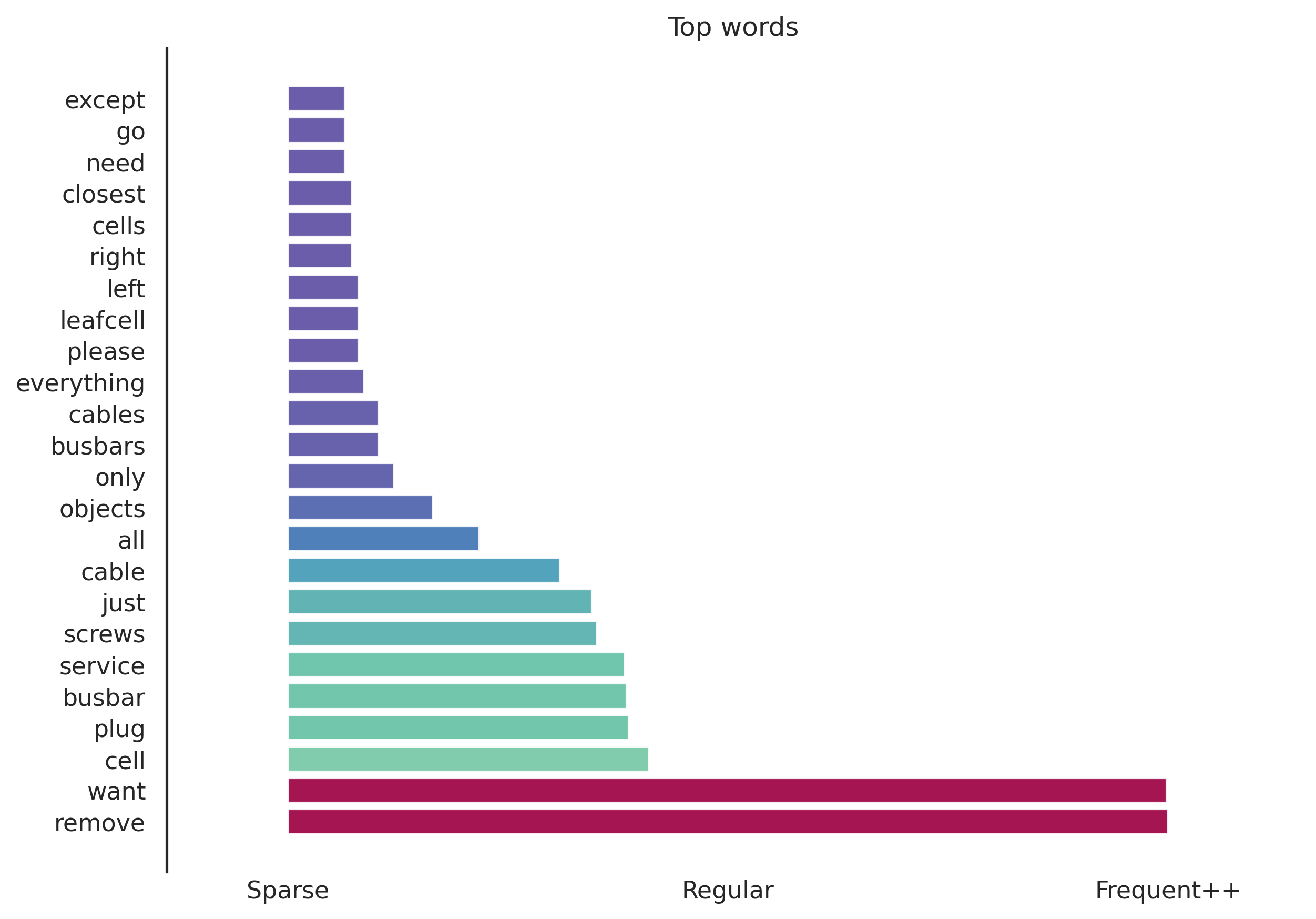}
    \caption{
    Token frequency across 600 operator requests. Bars are ordered from rare to common. Colours grade from purple (sparse) through teal to red (frequent). Red bars highlight the two dominant verbs (\textit{want}, \textit{remove}); the teal gradient marks object classes (\textit{LeafCell}, \textit{Service Plug}, \textit{Busbar}, \textit{Screw}, \textit{Cable}) following in prominence. At the same time, other less prominent words show in purple.}
    \label{fig:circle_words}
\end{figure*}

\section{Results}\label{sec5}
This section reports three things. First, a brief analysis of the language used in the operator requests to form the dataset is presented. Second, a quantitative study of the planning pipeline at various configurations is conducted, comparing the different ablations of the system. Third, the workload analysis and real-world assessment results.

For the pipeline quantitative study, we evaluate two targets: (i) \textit{Full Correctness} (the full removal set in correct order), and (ii) \textit{Next Object Correctness} (the first object only). For both targets, we report accuracy and time (average and median) per stage and ensemble size.

\subsection{Language Results}
We start with the request language because it sets the operating context for the planner. The instructions given by the operator shape intent and action classes, while the presence of specific object names sets the scope of the feasible plans. Figure \ref{fig:circle_words} summarises the most frequent terms and object references.

The dominant words appearing on the requests given by the operators were \textit{want} and \textit{remove}, which align with intent specification (want) and the disassembly task (remove). Object terms (\textit{leafcell}, \textit{service plug}, \textit{busbar}, \textit{screw}, \textit{cable}) appear high in the list, with \textit{leafcell} being the most mentioned object, and \textit{cable} the least.

Numerically, aggregating singular and plural mentions, the \textit{cable} is mentioned more than 90 times, \textit{screws} more than 100, \textit{service plugs} more than 120, \textit{busbars} more than 125, and \textit{leafcells} more than 135. Generic selectors such as \textit{all}, \textit{objects}, and \textit{everything} were used more than 70 times across the 600 requests. These frequencies describe the request mix that the ablations evaluate. These do not change the scoring rules but explain the prevalence of certain object combinations and the length of outputs in the test cases.

\subsection{Pipeline Results}\label{sec:pipeline_results}
We evaluated the pipeline on an operator-prepared fixed dataset of scenes and prompts, separate from the free-form “play” with non-expert participants. For each scene, we defined one or more admissible ground-truth removal sequences that satisfy disassembly precedence. The LLM received an intent prompt that could be a full or partial order, or just a target object; the only requirement was that the operator prompt reference at least one object in the current scene. The LLM was allowed to reorder objects and to insert mandatory prerequisites (e.g., removing screws or cables before a leafcell) that the user did not explicitly request.

Accuracy was computed with two objective metrics: \textbf{Full-sequence correctness:} the predicted ordered list exactly matches any admissible ground-truth sequence. \textbf{Next-object correctness:} the first predicted object is the correct next action, regardless of the rest of the list. Timing uses the same executions for both analyses. Therefore, average and median times are reported once per configuration and shared across the Full and Next evaluations.

\begin{table}[!h]
\centering
\footnotesize
\caption{Pipeline results from the same executions. Accuracy for \textit{Full} and \textit{Next} analyses, with shared average and median times per configuration.}
\label{tab:pipeline-results}
\begin{tabular}{clrr|rr}
\toprule
\multicolumn{2}{c}{} & \multicolumn{2}{c}{\textbf{Accuracy}} & \multicolumn{2}{c}{\textbf{Time (s)}} \\
\cmidrule(lr){3-4}\cmidrule(lr){5-6}
\textbf{LLMs} & \textbf{Stage} & \textbf{Full} & \textbf{Next} & \textbf{Avg} & \textbf{Med} \\
\midrule
1 & LLM      & 0.761 & 0.866 &  5.571 &  4.260 \\
\hdashline
3 & Ensemble & 0.787 & 0.863 & 21.934 & 19.250 \\
3 & Final    & 0.796 & 0.871 & 22.153 & 19.300 \\
\hdashline
6 & Ensemble & 0.816 & 0.888 & 32.504 & 28.935 \\
6 & Final    & 0.824 & 0.894 & 32.744 & 29.005 \\
\bottomrule
\end{tabular}
\end{table}
\vspace{-0.1em}

\begin{figure}[!h]
    \centering
    \includegraphics[width=1.0\linewidth]{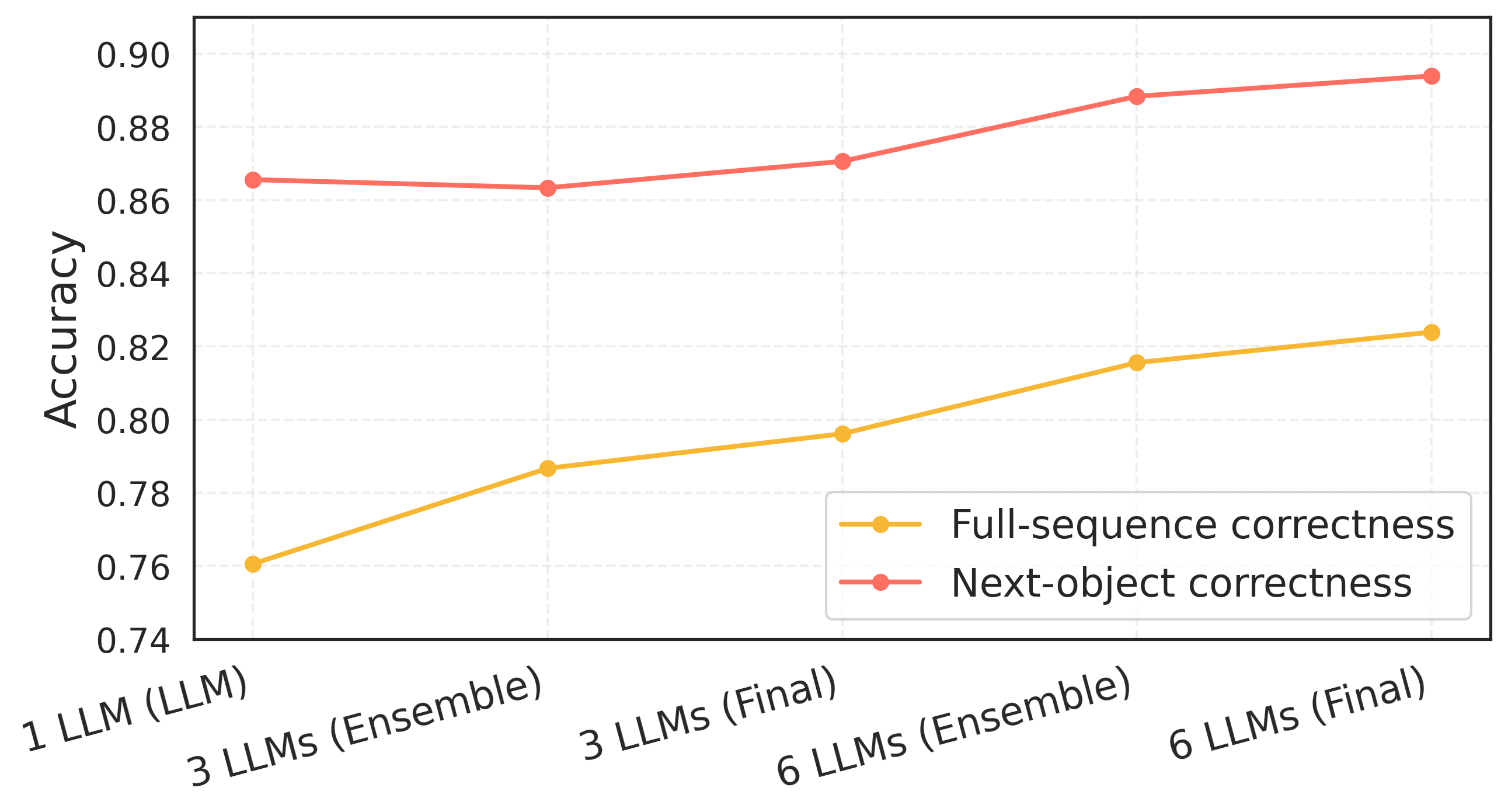}
    \caption{Accuracy across pipeline configurations for Full and Next-object evaluations.}
    \label{fig:accuracy_trend}
\end{figure}

On Table \ref{tab:pipeline-results}, the accuracies for the different configurations of \textit{1 LLM}, \textit{3 LLMs} ensemble + verifier (Ensemble), and ensemble + verifier + algorithmic verifier (Final), and \textit{6 LLMs} ensemble + verifier (Ensemble), and ensemble + verifier + algorithmic verifier (Final), are reported, for both \textit{full} correctness, and \textit{next object} correctness trials, accompanied by the respective time of their calculations. For a clearer observation and easier comparison of the accuracy gains, Figure \ref{fig:accuracy_trend} presents the trendline for both sets of trials across the various pipeline configurations. 

\begin{figure}[!h]
    \centering
    \includegraphics[width=1.0\linewidth]{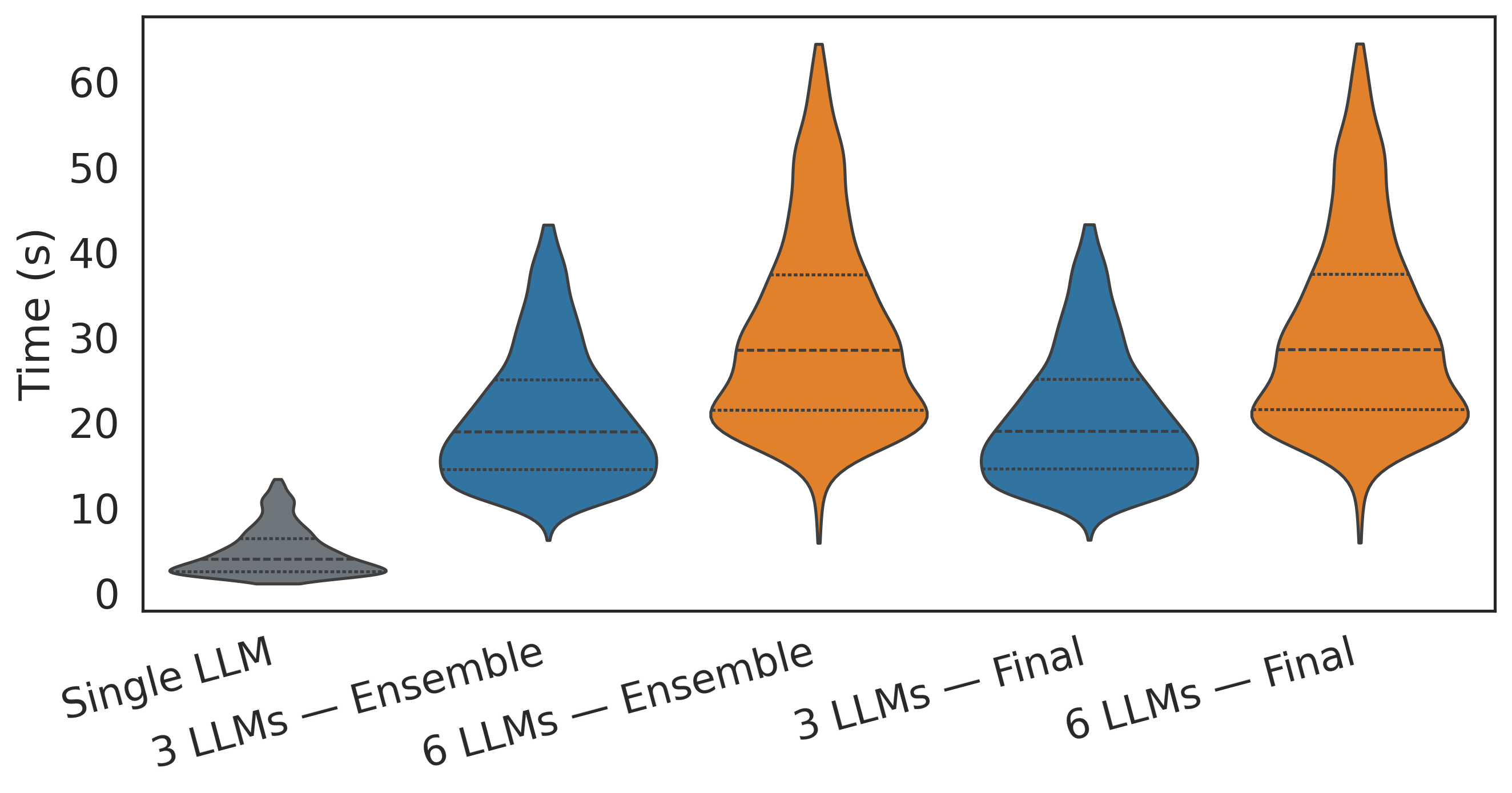}
    \caption{End-to-end per-request runtime cleaned  Interquartile range distribution with embedded boxplot.}
    \label{fig:violin-clean-full}
\end{figure}

\begin{figure}[!h]
    \centering
    \includegraphics[width=1.0\linewidth]{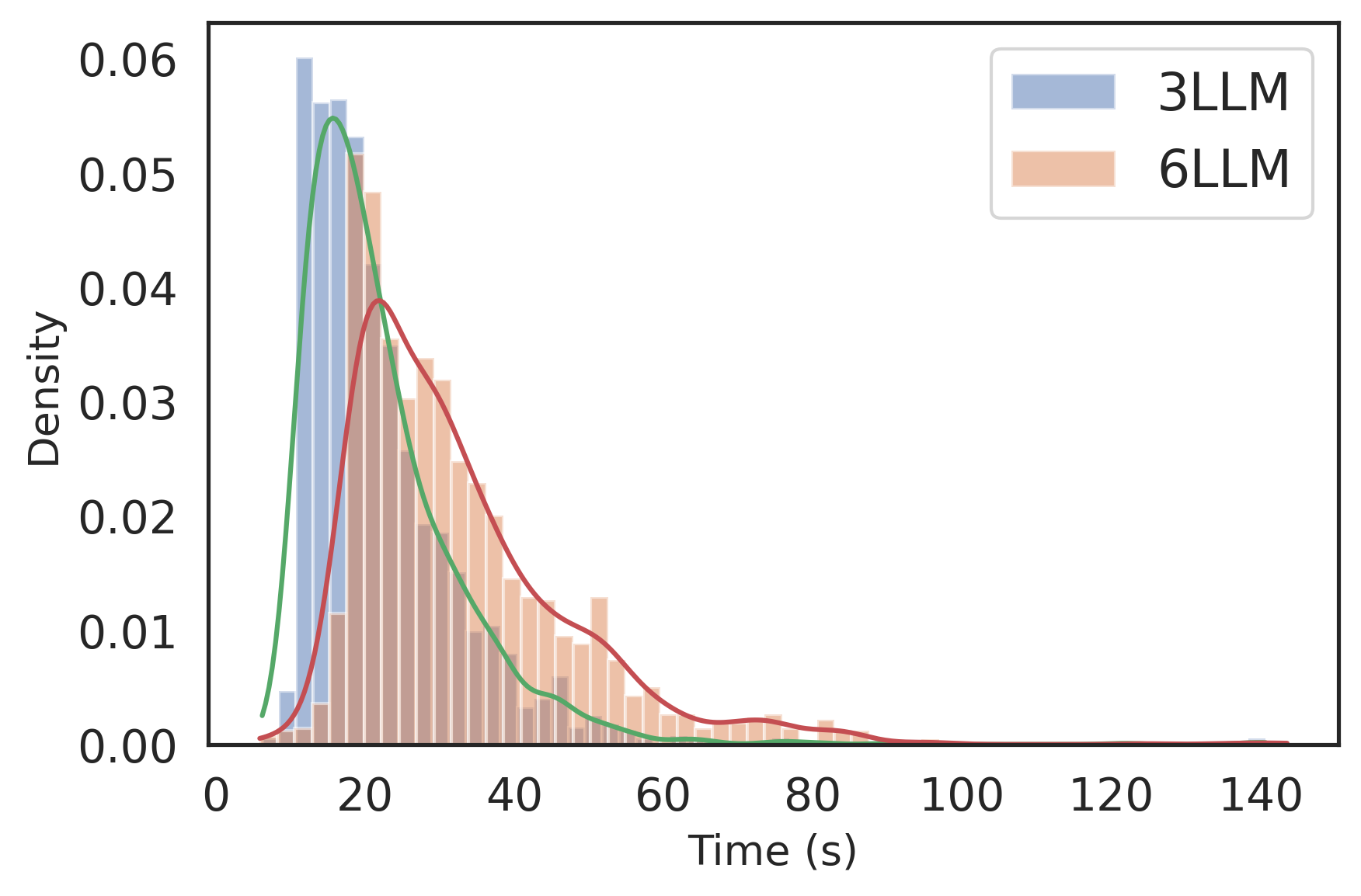}
    \caption{Final-stage runtime distribution (9,000 runs). Density-normalised histograms with KDE for 3LLM and 6LLM pipelines; applies to both Full and Next object evaluations. Effective bandwidths are $\approx$2.45 s (3 LLM) and $\approx$3.36 s (6 LLM).}
    \label{fig:time-hist-full}
\end{figure}

Figure \ref{fig:violin-clean-full} displays a violin plot for observation of the distribution of time per request across the dataset, with Figure \ref{fig:time-hist-full} complementing this, by reporting the distribution of final stage times for the 9000 runs with density values for each of the time bandwidths.

%%%%%%%%%%%%%%%%%%%%%%%%%%%%%%%%%%%%%%%
%Violin and histogram
%%%%%%%%%%%%%%%%%%%%%%%%%%%%%%%%%%%%%%%

% \vspace{-1.2em}

%%%%%%%%%%%%%%%%%%%%%%%%%%%%%%%%%%%%%%%
% END
%%%%%%%%%%%%%%%%%%%%%%%%%%%%%%%%%%%%%%%

\begin{figure*}[!h]
    \centering
    \begin{minipage}[t]{0.49\linewidth}
        \centering
        \includegraphics[width=\linewidth]{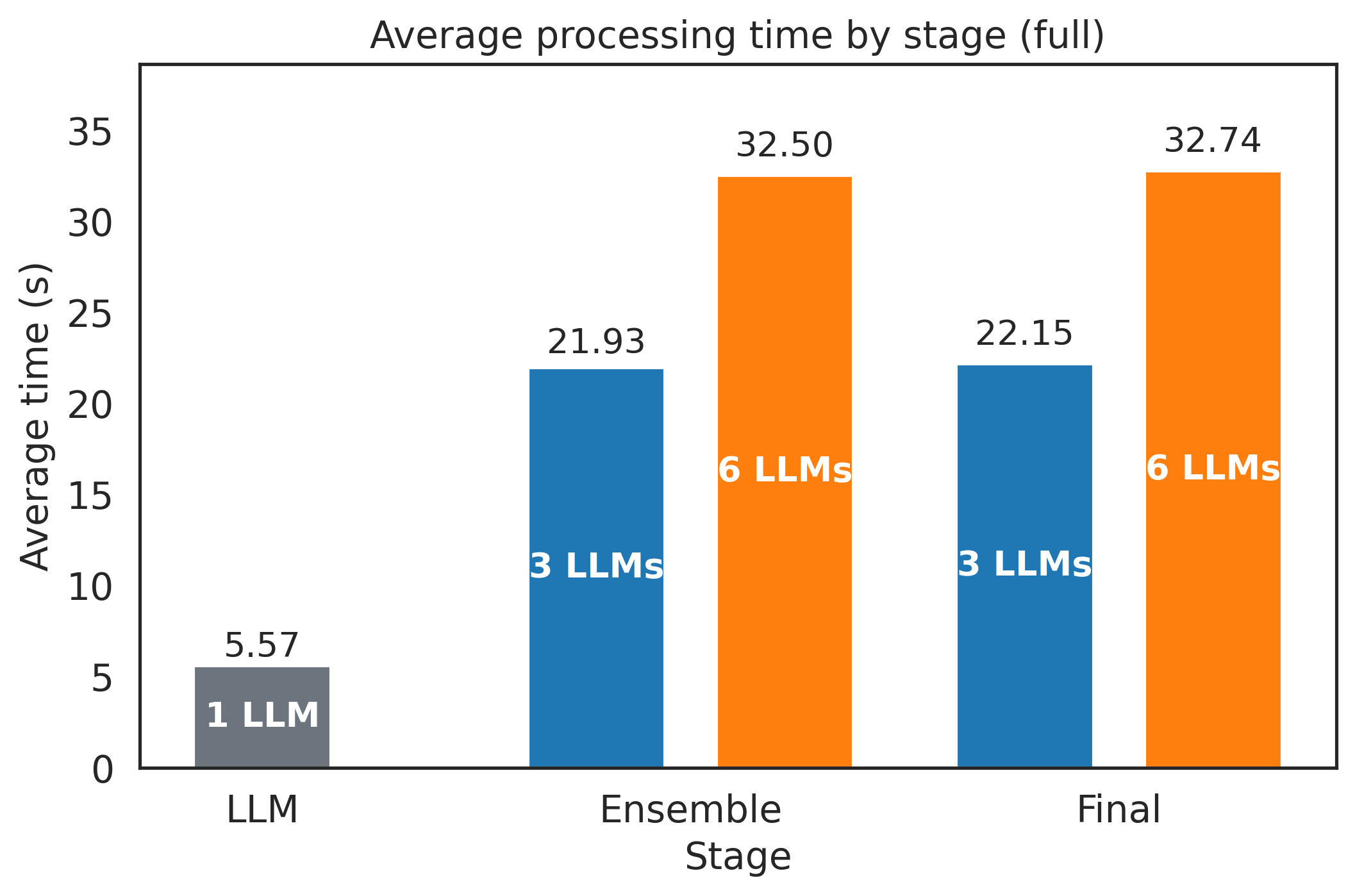}\\[-2pt]
        \footnotesize Average time by stage
    \end{minipage}
    \hfill
    \begin{minipage}[t]{0.49\linewidth}
        \centering
        \includegraphics[width=\linewidth]{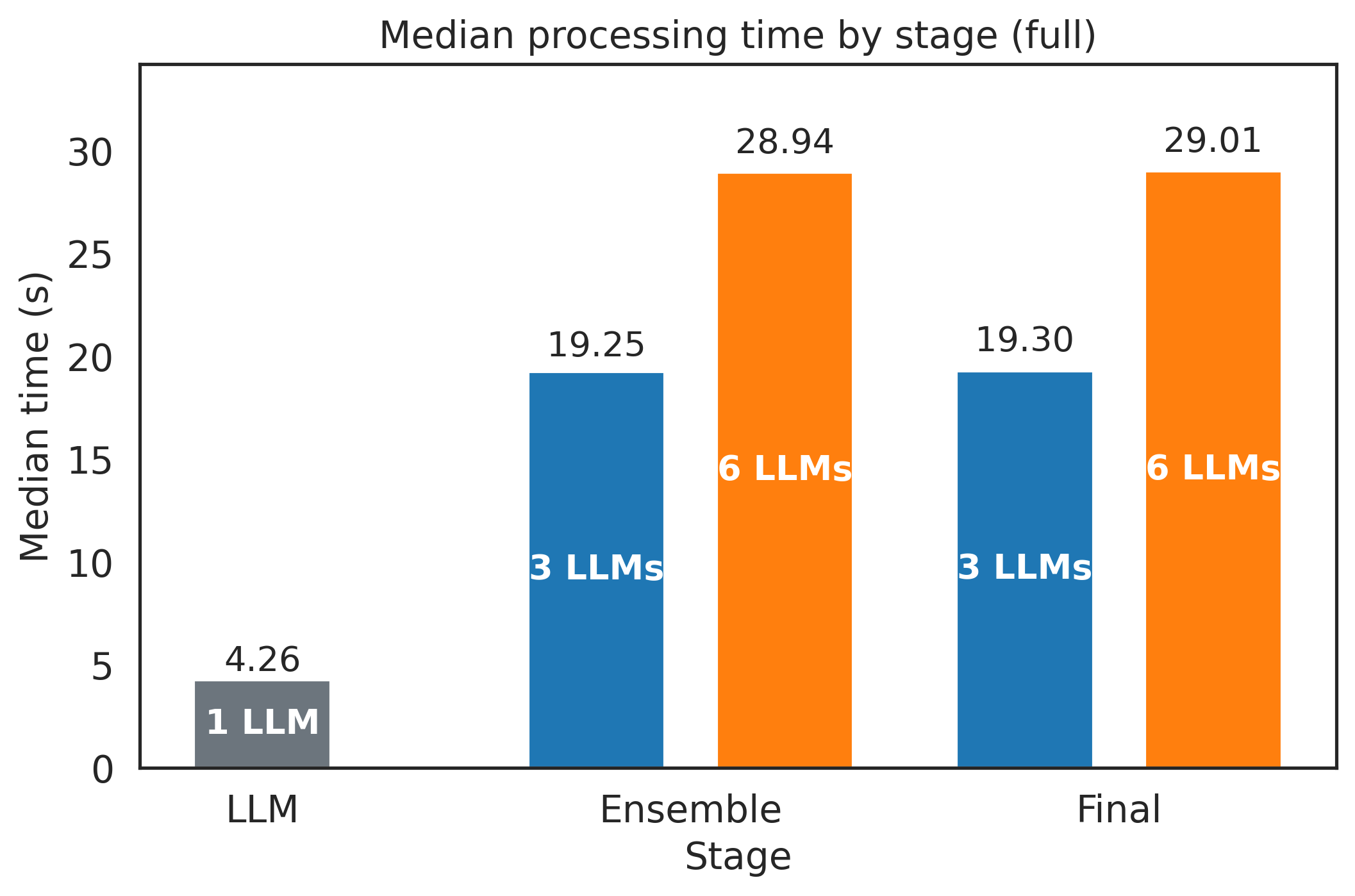}\\[-2pt]
        \footnotesize Median time by stage
    \end{minipage}
    \caption{Processing time by stage. Left: average time; right: median time. The $x$-axis lists stages (\textit{LLM}, \textit{Ensemble}, \textit{Final}). Within each stage, bars compare configurations: single-model for the 1 LLM configuration, and 3-model vs 6-model for \textit{Ensemble} and \textit{Final}. The $y$-axis is seconds.}
    \label{fig:time-bars}
\end{figure*}

\begin{figure*}[!h]
    \centering
    \begin{minipage}[t]{0.47\linewidth}
        \centering
        \includegraphics[width=\linewidth]{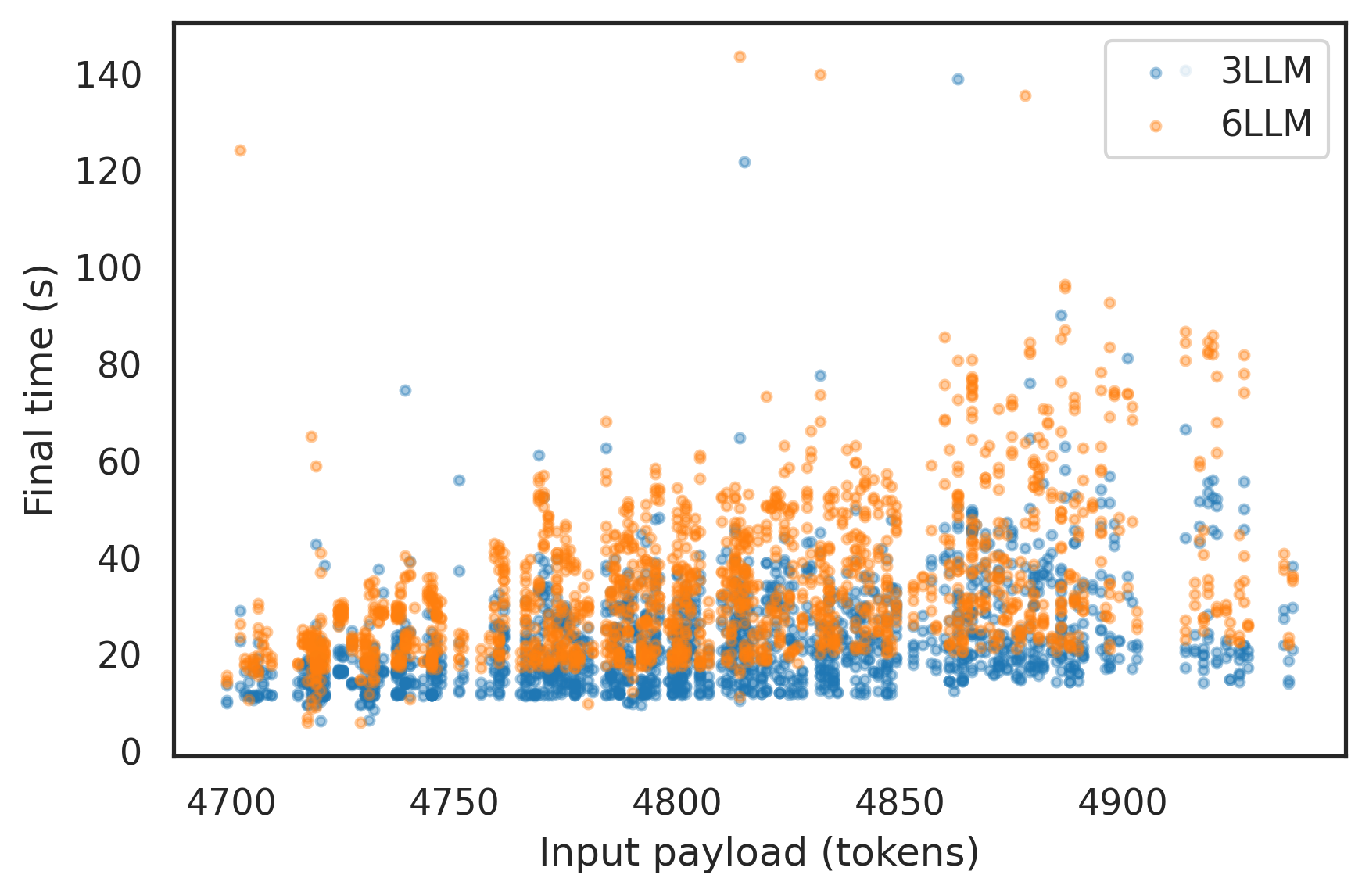}\\[-2pt]
        \footnotesize Input tokens vs final-stage latency (Final, full)
    \end{minipage}
    \hfill
    \begin{minipage}[t]{0.47\linewidth}
        \centering
        \includegraphics[width=\linewidth]{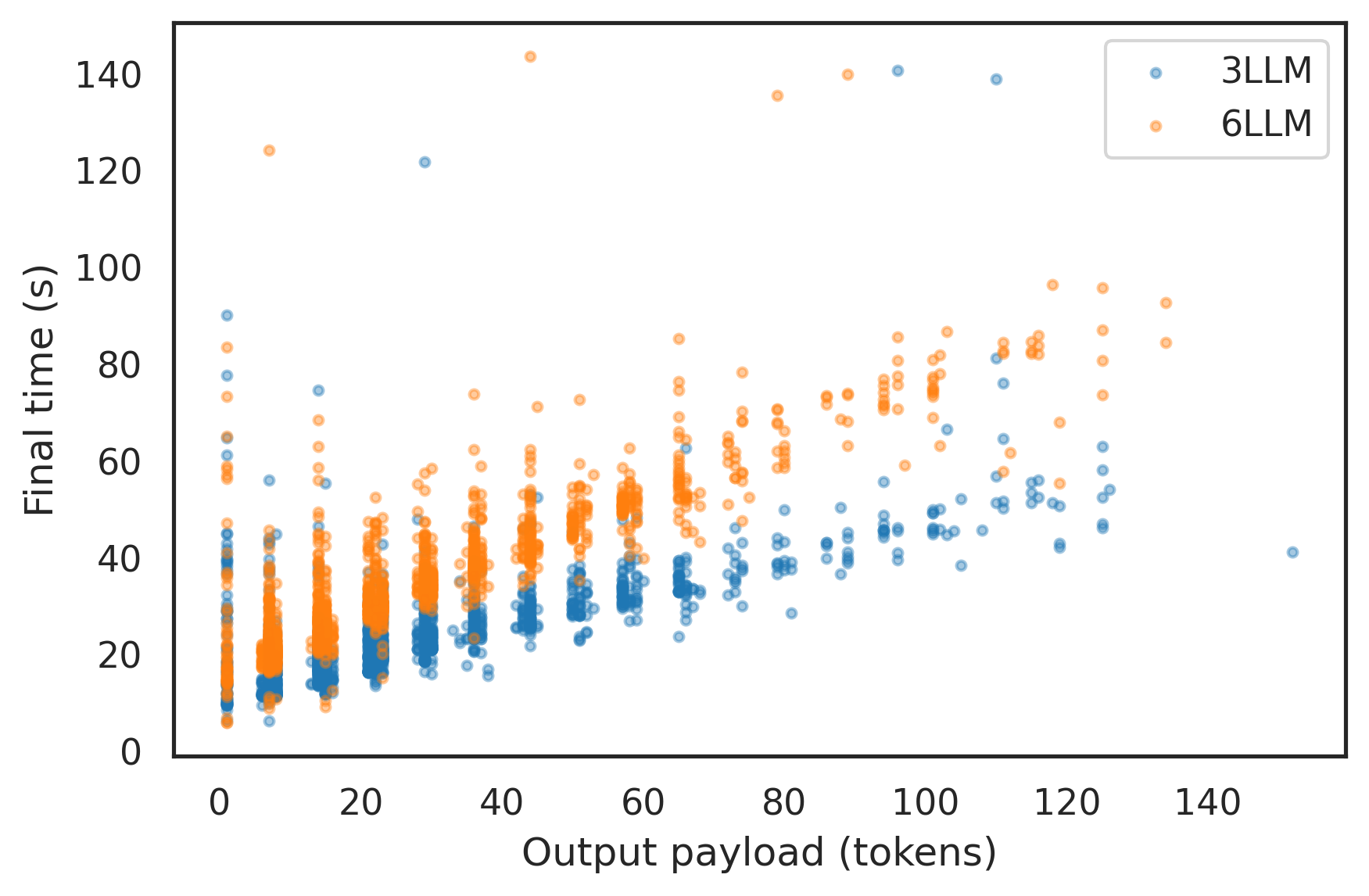}\\[-2pt]
        \footnotesize Output tokens vs final-stage latency (Final, full)
    \end{minipage}
    \caption{Payload size vs latency until end of the \textit{Final} ensemble stage. Left: input token count; right: output token count. Each point is a single request from the same executions as Table~\ref{tab:pipeline-results}. Tokens are counted by the model tokeniser; latency is the time of the \textit{Final} stage. Separate series are shown for the 3-model and 6-model ensembles.}
    \label{fig:payload-vs-latency}
\end{figure*}

Figure \ref{fig:time-bars} reports per-stage processing times for the full evaluation. The left panel displays the mean time per request, while the right panel shows the median time per request. The x-axis lists stages (LLM, Ensemble, Final). Within each stage, bars correspond to the 1LLM baseline for LLM and to the 3LLM and 6LLM configurations for Ensemble and Final. The y-axis is in seconds, measured at the last word output time.

Figure \ref{fig:payload-vs-latency} plots payload size (tokens used) against final-stage latency for the full evaluation. The left panel uses input token counts; the right panel uses output token counts. Each point is a single request from the same executions as Table \ref{tab:pipeline-results}. The model tokeniser measures tokens; latency is the time recorded for the Final stage. The separate series indicate the 3LLM and 6LLM ensembles.

Table \ref{tab:stage_pvalues} lists p-values for pairwise stage comparisons (LLM vs Ensemble, LLM vs Final, Ensemble vs Final) under the Full (all objects) and Next (next object) settings, for 3LLM and 6LLM configurations. Entries with p less than 0.05 are marked with a checkmark. We used two-sided paired t-tests on per-prompt accuracy to compare stages, leveraging the fact that the same prompts were evaluated at each stage to control for between-prompt variability. This test assumes independence across prompts and approximate normality of the paired differences, with large N (1800). The table reports p-values only.\

Note that prompts could be partial; the LLM often inserted necessary prerequisites (e.g., screws or cables) before the requested target to satisfy precedence while still matching operator intent under our scoring rules.
\vspace{-0.5em}

%%%%%%%%%%%%%%%%%%%%%%%%%%%%%%%%%%%%%%%
% Table of p-values (paired t-test)
%%%%%%%%%%%%%%%%%%%%%%%%%%%%%%%%%%%%%%%
\begin{table*}[!h]
\centering
\footnotesize
\caption{Stage comparisons using \textbf{paired t-tests} on matched 0/1 correctness (two-sided). A checkmark indicates $p<0.05$.}
\label{tab:stage_pvalues}
\begin{tabular}{llcc|cc}
\toprule
\textbf{Comparison} & \textbf{Setting} & \textbf{3LLM $p$} & \textbf{Sig.@0.05} & \textbf{6LLM $p$} & \textbf{Sig.@0.05} \\
\midrule
\multirow{2}{*}{LLM vs Ensemble}
  & Full  & \textbf{$2.19\times 10^{-7}$} & \checkmark & \textbf{$1.57\times 10^{-9}$} & \checkmark \\
  & Next  & \textbf{$2.90\times 10^{-2}$} & \checkmark & \textbf{$1.13\times 10^{-3}$} & \checkmark \\
\midrule
\multirow{2}{*}{LLM vs Final}
  & Full  & \textbf{$6.61\times 10^{-10}$} & \checkmark & \textbf{$1.19\times 10^{-11}$} & \checkmark \\
  & Next  & \textbf{$1.45\times 10^{-3}$} & \checkmark & \textbf{$7.70\times 10^{-5}$} & \checkmark \\
\midrule
\multirow{2}{*}{Ensemble vs Final}
  & Full  & \textbf{$3.61\times 10^{-5}$} & \checkmark & \textbf{$1.74\times 10^{-3}$} & \checkmark \\
  & Next  & \textbf{$7.78\times 10^{-4}$} & \checkmark & \textbf{$2.68\times 10^{-3}$} & \checkmark \\
\bottomrule
\end{tabular}
\end{table*}
%\vspace{-1.2em}

\subsection{Real-world Pilot Results}
\label{sec:real_world_demo_results}

Seven volunteers (\textit{n}=7) completed one session each. The study protocol, including all operator-in-the-loop experiments, was approved by the University of Birmingham Ethics Committee under project ERN\_3337. Each participant issued two intents (one per arm), totalling 14 intended actions. All disassembly actions were successfully planned as intended by the participants through the framework. Success was achieved for every participant in both cases in 14/14 actions.

Table \ref{tab:time_summary} reports the mean and median completion time for the participants using the pipeline for the control of the robots. A 300s manual baseline serves as a contextual reference from an expert operator, performing the task without LLM-assisted planning, with which to compare this data.

%\vspace{-1.2em}
\begin{table}[!h]
\centering
\footnotesize
\caption{Completion time (pipeline condition). Values are Mean $\pm$ SD and Median, in seconds.}
\label{tab:time_summary}
\begin{tabular}{lcc}
\toprule
Metric & Mean $\pm$ SD (s) & Median (s) \\
\midrule
Time to action & 197.86 $\pm$ 19.61 & 195.00 \\
\bottomrule
\end{tabular}
\end{table}
%\vspace{-1.2em}

Table \ref{tab:tlx_global_summary} summarises the global NASA-TLX workload (0-100). Used to report the overall perceived workload for the session.

%\vspace{-0.8em}
\begin{table}[!h]
\centering
\footnotesize
\caption{Global NASA-TLX workload (0-100). Values are Mean $\pm$ SD and Median.}
\label{tab:tlx_global_summary}
\begin{tabular}{lcc}
\toprule
Metric & Mean $\pm$ SD & Median \\
\midrule
NASA-TLX (Global) & 14.95 $\pm$ 11.80 & 9.67 \\
\bottomrule
\end{tabular}
\end{table}
%\vspace{-0.8em}

Table \ref{tab:tlx_subscales_summary} reports mean ± SD and median for each NASA-TLX subscale (0-100): Mental Demand, Physical Demand, Temporal Demand, Performance, Effort, and Frustration.

\begin{table}[!h]
\centering
\footnotesize
\caption{NASA-TLX subscales (0-100): Mean $\pm$ SD and Median across participants.}
\label{tab:tlx_subscales_summary}
\begin{tabular}{lcc}
\toprule
Subscale & Mean $\pm$ SD & Median \\
\midrule
Mental Demand    & 10.00 $\pm$ 5.77 & 10.00 \\
Physical Demand  &  7.86 $\pm$ 5.67 &  5.00 \\
Temporal Demand  &  7.86 $\pm$ 5.67 &  5.00 \\
Performance      & 27.86 $\pm$ 34.03 & 10.00 \\
Effort           & 10.00 $\pm$ 5.00 & 10.00 \\
Frustration      & 10.00 $\pm$ 6.46 &  5.00 \\
\bottomrule
\end{tabular}
\end{table}
%\vspace{-1.2em}

\newpage
.
\newpage

\section{Discussion}

This work tested intent-driven ensemble LLM planners, performing a case study on EV-battery component disassembly using two UR10e arms. However, it's capable of adapting to different environments, robots, and scenarios.

\subsection{Model selection}
As the model selection and preliminary experiments show, the most important characteristic when choosing among low parameter count models was their instruction adherence capability. The most common errors found in models that didn't adhere to the instructions were making formatting errors or being overly verbose in explaining their conclusions before providing the list of components, even though the instructions explicitly requested only the list as output. Qwen3-32B (reasoning off) provided the best responses, although its speed remained above 20 seconds on average.

\subsection{Pipeline correctness}
Across the 9000 runs (Table \ref{tab:pipeline-results}), \textbf{full object} correctness rose from 0.761 (single model) to 0.816 (6-model+verifier) and to 0.824 with the final algorithmic filter. That is a +6.3 percentage-point absolute gain over the single model or $\approx$+8.3\% relative gain. Accuracy increases with ensemble size (Figure~\ref{fig:accuracy_trend}). The deterministic filter adds small but consistent increments (0.9 pp for 3LLM and 0.8 pp for 6LLM) while removing hallucinated objects at nearly zero time cost. Paired t-tests on matched per-prompt accuracy (two-sided) confirm significant improvements from single-model to Ensemble and to the Final all integrated framework (Full: 3LLM \(p=2.19\times10^{-7}\), 6LLM \(p=1.57\times10^{-9}\); LLM \(\rightarrow\) Final: 3LLM \(p=6.61\times10^{-10}\), 6LLM \(p=1.19\times10^{-11}\)). Ensemble \(\rightarrow\) Final yields small but significant gains (Full: 3LLM \(p=3.61\times10^{-5}\), 6LLM \(p=1.74\times10^{-3}\)); see Table~\ref{tab:stage_pvalues}.

For \textbf{next-object} correctness, the single model gives a high accuracy (0.866). The 3LLM configuration achieves 0.863 (Ensemble) and 0.871 (Final), while the 6LLM configuration reaches 0.888 (Ensemble) and 0.894 (Final), indicating an increasing trend with a slight dip at 3LLM before rising at 6LLM. Paired t-tests show significant gains over the single model (3LLM: LLM \(\rightarrow\) Ensemble \(p=2.90\times10^{-2}\), LLM \(\rightarrow\) Final \(p=1.45\times10^{-3}\); 6LLM: \(p=1.13\times10^{-3}\) and \(7.70\times10^{-5}\)). Ensemble \(\rightarrow\) Final adds further small but significant improvements (3LLM \(p=7.78\times10^{-4}\); 6LLM \(p=2.68\times10^{-3}\)); Table~\ref{tab:stage_pvalues}.

Latency increases with ensemble size: mean end-to-end time is 5.57s (single model), 21.93–22.15s (3-model, ±filter), and 32.50–32.74s (6-model, ±filter) (Table \ref{tab:pipeline-results}). Median times show the same pattern (4.26s → 19.25–19.30s → 28.94–29.01s). The final filter adds approximately 0.2 seconds on average, which is negligible compared to ensemble sampling and verification. Distributions are right-tailed (Fig. \ref{fig:violin-clean-full}) and widen with ensemble size, with the final-stage histograms (Fig. \ref{fig:time-hist-full}) showing the bandwidth increasing from $\approx$2.45s (3-model) to $\approx$3.36s (6-model). The whole pipeline’s mean latency scales to approximately 5.9 times the single-model baseline (+488\% relative), yielding a respectable and expected compute-accuracy trade-off. For a planning pipeline, although these numbers are slower compared to pre-planned methods or distance-based planning, they are suitable for tasks that require some adaptability and more intelligent, dynamic planning capabilities under different conditions. Token–latency Figures \ref{fig:payload-vs-latency} suggest a weak positive correlation between payload and final-stage time, with most of the latency changes in the spread dominated by ensemble size rather than token counts.

The measured accuracy lift provided by the deterministic filter is small but consistent across both metrics. It comes with a mean time overhead of approximately $\mathbf{0.20}\,\mathrm{s}$ (Fig. \ref{fig:time-bars}), acting as a very effective low-cost last line of defence against hallucinations. For our framework the scene included five component classes (leafcell, busbar, cable, service plug, screw, Figure \ref{fig:image_analysis}) detected by YOLOv8, although the LLM model was never given the explicit labels it was able to properly adapt with the provided instructions, to the components of the disassembly task, making a strong case for the generalizability of the pipeline, which cared more about the data format and structure consistency (with no model working in preliminary tests, with unstructured data).

The bar plot of prompt tokens (Fig. \ref{fig:circle_words}) shows that operator intents are dominated by "want" and "remove", due to the disassembly nature of the trials, with object mentions led by "leafcell" and "busbar". Generic selectors, such as "all/objects/everything", appear frequently, explaining plans that contained a high number of objects and why precedence checks matter in disassembly tasks.

\subsection{Human Pilot Experiment}
The results of the pilot suggest that a language-level variable autonomy mechanism with verification maintains low operator workload while preserving user control. 

First, all 14 intents were executed only after participants reviewed and approved the candidate lists. The final execution remained operator-controlled. 
Second, perceived workload was low \cite{nikulin2019nasa, grier2015high}(NASA–TLX mean $14.95$, median $9.67$) after a brief single training session (Tables~\ref{tab:time_summary}–\ref{tab:tlx_subscales_summary}). The UI only displayed the detected objects by YOLO, in the form of a 2D image and a 3D object display. Participants did not tune any low-level motion or grasp parameters; they only issued intents and approved an ordered list. This likely reduced cognitive load. Third, time-to-action averaged 197.86\,s (median 195\,s), lower than a contextual manual reference of \(\approx 300\,\text{s}\), indicating that plan synthesis plus checking added little overhead relative to manual planning.

The verifier LLM and the final algorithmic filter removed format and consistency errors before execution, limiting cognitive effort to intent expression and a single "Go" confirmation. The sample is small and lacks an interface control, but the pattern aligns with variable autonomy: the autonomy proposes and checks plans; the operator authorises execution.

\section{Conclusion}
This work presented an intent-driven planning pipeline for multi-robot collaboration and disassembly that integrates perception-to-text scene encoding, an LLM ensemble sampled from a single checkpoint, a format/priority LLM verifier, and a final deterministic consistency filter wired end-to-end. 

An ensemble-with-verification architecture improves multi-robot disassembly planning over a single LLM. On 200 real scenes and 600 intents, full-sequence accuracy rises from 0.761 to 0.824 with a 6LLM ensemble plus an LLM verifier and a final deterministic filter; next-object accuracy rises from 0.866 to 0.894. Under \emph{paired t-tests}, gains are significant, and the deterministic filter provides additional small, significant, safety-oriented improvements at negligible time cost. Latency scales with ensemble size (\(\approx 5.9\times\) the single-model baseline), reflecting the standard accuracy–compute trade-off.

In real-world use, non-experts could issue intents that were executed as intended after list review, with a low perceived workload ($\approx$ 15) and shorter time-to-action than with a contextual manual reference (300s). Time-to-action was faster than the manual baseline by roughly one-third on average ($\approx34\%$ faster), with the median about 35\% faster, and run-to-run variability $\pm10\%$ relative to the mean.

The LLM ensemble, equipped with explicit verification and data consistency checks, results in a viable approach to injecting reliability and adaptability into intent-driven planning for collaborative disassembly.

\subsection{Limitations and Future Work}
Our ensemble improves plan correctness but shows the expected accuracy--latency trade-off: accuracy gains taper while mean latency increases from 5.57\,s (single model) to 32.74\,s (6-LLM) on the same prompts. Future work will incorporate dynamic computation, including early-exit voting when candidates agree, as well as input-complexity gating between 1/3/6 models. We also plan to cache and reuse verified partial plans for recurring intents (when possible).

Verification is text-only today (format/precedence rules plus a data consistency filter). This prevents many errors, but cannot enforce geometric feasibility. We will add geometry-aware checks (reachability, collisions, graspability) to reject impossible steps and to provide minimal counter examples that guide operator edits. The pipeline currently assumes reliable detections, but is highly dependent on the point cloud and the YOLOv8 model weights. To improve the detection rates, we will incorporate multi-view fusion and uncertainty-aware perception to enhance its accuracy. A pre-grasp re-verification stage with live sensing and per-action risk scores (e.g. hazard proximity or collision) can enable safe local reordering within the stated intent.

Diversity in our results stems from one checkpoint with different seeds. Mixing model families and prompt variants may outperform the same-checkpoint ensembles at a similar computational cost, which is also worth investigating as part of future work. Although the verifier prompt is domain-agnostic, we evaluated one domain with five component classes. We will test other assembly/disassembly settings (e-waste, other types of end-of-life product dismantling, nuclear decommissioning). Another area for deeper study in future work, is expanding the human user study to include larger numbers of participants, and exploring if and how human-perceived workload varies across different types of tasks and environments.

\section*{Conflict of Interest Statement}
The authors declare no conflict of interest.

\section*{Author Contributions}
Conceptualisation, C.E. and C.A.C.; methodology, C.E. and C.A.C.; software, C.E. and C.A.C.; validation, C.E., C.A.C. and M.C.; investigation, C.E., C.A.C. and A.R.; data curation, C.E. and C.A.C.; writing—original draft preparation, C.E., C.A.C; writing—review and editing, C.E., C.A.C., R.S., M.C. and A.R.; visualisation, C.E. and C.A.C; supervision, A.R. and R.S.; project administration, A.R., R.S.; funding acquisition, A.R. and R.S. All authors have read and agreed to the published version of the manuscript.

\bibliographystyle{IEEEtran}
\bibliography{main}

\end{document}